%% file: acl_latex.tex
\pdfoutput=1

\documentclass[11pt]{article}

\usepackage[]{acl}
\usepackage{enumitem}

\usepackage{times}
\usepackage{latexsym}
\usepackage{graphicx}
\usepackage{booktabs}
\usepackage{svg}
\usepackage{xcolor}
\usepackage{csquotes}
\usepackage{amsmath}
\usepackage{amssymb}
\usepackage{cleveref}
\usepackage{hyperref}
\usepackage{multirow} 

\definecolor{light-gray}{gray}{0.95}

\usepackage[T1]{fontenc}

\usepackage[utf8]{inputenc}

\usepackage{microtype}

\usepackage{inconsolata}
\usepackage{soul}

\title{Code Prompting Elicits Conditional Reasoning Abilities\\in Text+Code LLMs}

\author{Haritz Puerto$^{1}$, Martin Tutek$^{2}$\thanks{Work done while author was at TU Darmstadt.}, Somak Aditya$^{3}$, Xiaodan Zhu$^{1,4}$, Iryna Gurevych$^{1}$\\
$^{1}$Ubiquitous Knowledge Processing Lab (UKP Lab),\\
TU Darmstadt and Hessian Center for AI (hessian.AI) \\
$^{2}$Technion -- IIT, $^{3}$IIT Kharagpur, $^{4}$Queen's University\\ \href{https://www.ukp.tu-darmstadt.de}{https://www.ukp.tu-darmstadt.de}
}

\begin{document}
\maketitle
\begin{abstract}
\input{latex/sections/abstract}
\end{abstract}

\section{Introduction}
\input{latex/sections/1-introduction}

\begin{figure*}[t]
\centering
\includegraphics[width=0.9\textwidth]{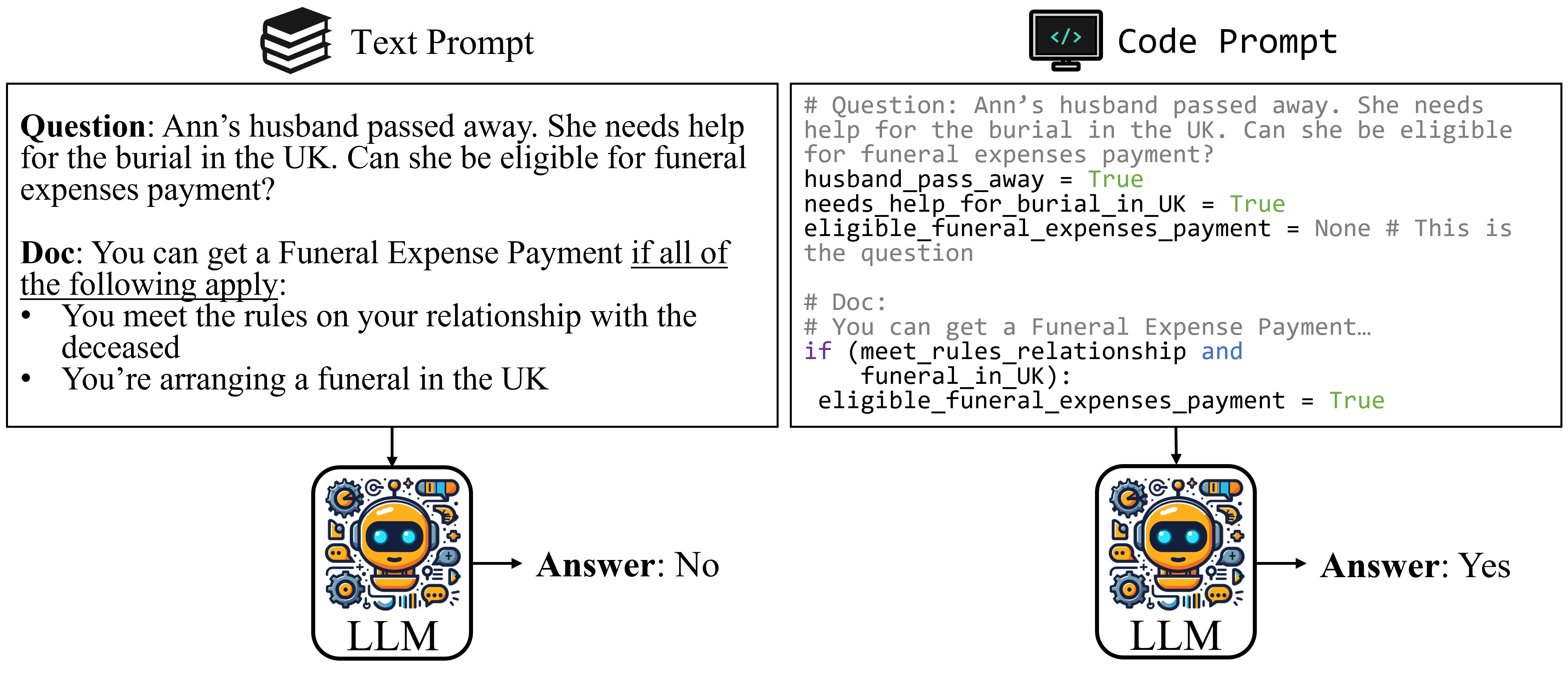}
\caption{\texttt{Code prompting} converts natural language descriptions into code to be solved with a large language model. The figure shows a transformed instance from the ConditionalQA dataset.}
\label{fig:code_prompt}
\end{figure*}

\section{Background and Related Work}
\input{latex/sections/2-rw}

\section{Code Prompting}
\input{latex/sections/3-model}
\section{Experimental Setup}
\input{latex/sections/4-setup}

\input{latex/new_tables/overall_results}
\section{Experiments}
\input{latex/sections/5-experiments}

\section{Human Analysis of the Generated Code}
\label{appendix:human_eval}
\input{latex/sections/appendix/human_evaluation}

\section{Conclusions and Future Work}
\input{latex/sections/6-conclusions}

\section*{Limitations}
\input{latex/sections/limitations}

\section*{Ethics and Broader Impact Statement}
\input{latex/sections/ethics}

\section*{Acknowledgements}
This work has been funded by the German Research Foundation (DFG) as part of the UKP-SQuARE project (grant GU 798/29-1), by the German Federal Ministry of Education and Research and the Hessian Ministry of Higher Education, Research, Science and the Arts within their joint support of the National Research Center for Applied Cybersecurity ATHENE, and by the European Union (ERC, InterText, 101054961). Views and opinions expressed are, however, those of the author(s) only and do not necessarily reflect those of the European Union or the European Research Council. Neither the European Union nor the granting authority can be held responsible for them. Somak Aditya acknowledges travel support from the European Union’s Horizon 2020 research and innovation program under grant agreement No 951847.

We gratefully acknowledge the support of Microsoft with a grant for access to OpenAI GPT models via the Azure cloud (Accelerate Foundation Model Academic Research).

Lastly, we thank Max Glockner, Jonathan Tonglet, Sheng Lu, and the anonymous reviewers for their insightful comments and suggestions on a prior draft of this paper.

\bibliography{anthology,cond}

\appendix

\section{List of Appendices}
We start the appendices with details on the datasets and prompts (\Cref{appendix:datasets} and \ref{appendix:prompts}). Then, we provide additional details on the features of the generated code, the setup of the LLMs, and their costs (\Cref{appendix:coding_features},  \ref{appendix:llm_setup}, and \ref{appendix:costs}). Subsequently, we perform experiments on code-only and text-only LLMs to show that code prompting should only be used on text+code LLMs (\Cref{appendix:codellm} and \ref{appendix:textllm}). Then, we show experiments on small LLMs (\Cref{appendix:phi2}), ablations on local LLMs (\Cref{appendix:ablations_local_llms},  \ref{appendix:demonstrations_local_llms}, \ref{appendix:variable_tracking}, and \ref{appendix:memory_errors_local_llms}. We conclude with the confusion matrices for the main experiments \Cref{appendix:confusion_matrix}, prompt examples \ref{appendix:atomic_statements}, \ref{appendix:code_ablations}, and \ref{appendix:prompt_examples}.

\section{Datasets}
\label{appendix:datasets}
\input{latex/sections/appendix/datasets}

\section{Prompt Formulation}
\label{appendix:prompts}
\input{latex/sections/appendix/prompts}

\section{Coding Features}
\label{appendix:coding_features}
\input{latex/sections/appendix/coding_features}

\section{LLM Setup}
\label{appendix:llm_setup}
\input{latex/sections/appendix/llm_setup}

\section{Costs}
\label{appendix:costs}
\input{latex/sections/appendix/costs}

\section{Code-only LLMs}
\label{appendix:codellm}
\input{latex/sections/appendix/codellm}

\section{Text-only LLMs}
\label{appendix:textllm}
\input{latex/sections/appendix/textllm}

\section{Results on Small LMs with Short Context Window}
\label{appendix:phi2}
\input{latex/sections/appendix/phi2_results}

\section{Ablations on Local LLMs}
\label{appendix:ablations_local_llms}
\input{latex/sections/appendix/ablations_local_llms}

\section{Number of Demonstrations on Local LLMs}
\label{appendix:demonstrations_local_llms}
\input{latex/sections/appendix/demonstrations_local_llms}

\section{Variable Tracking Setup}
\label{appendix:variable_tracking}
\input{latex/sections/appendix/variable-tracking}

\section{Memory Errors on Local LLMs}
\label{appendix:memory_errors_local_llms}
\input{latex/sections/appendix/memory_errors_local_llms}



\section{Confusion Matrices}
\label{appendix:confusion_matrix}
\input{latex/sections/appendix/confusion_matrix}

\section{Atomic Statements}
\label{appendix:atomic_statements}
\input{latex/sections/appendix/atomic_statements}

\section{Examples of Code Ablations}
\label{appendix:code_ablations}
\input{latex/sections/appendix/code_ablations}

\input{latex/sections/appendix/code_ablations_tables}
\input{latex/sections/appendix/variable_tracking_example}

\section{Prompt Examples}
\label{appendix:prompt_examples}
\input{latex/sections/appendix/prompt_examples}

\end{document}

%% file: latex/sections/abstract.tex
Reasoning is a fundamental component of language understanding. 
Recent prompting techniques, such as \textit{chain of thought}, have consistently improved LLMs' performance on various reasoning tasks. 
Nevertheless, there is still little understanding of what triggers reasoning abilities in LLMs in the inference stage.
In this paper, we investigate the effect of the \textit{input representation} on the reasoning abilities of LLMs. We hypothesize that representing natural language tasks as code can enhance specific reasoning abilities such as entity tracking or logical reasoning. To study this, we propose \textit{code prompting}, a methodology we operationalize as a chain of prompts that transforms a natural language problem into code and \textit{directly} prompts the LLM using the generated code \textit{without} resorting to external code execution.
We find that code prompting exhibits a high-performance boost for multiple LLMs (up to $22.52$ percentage points on GPT 3.5, $7.75$ on Mixtral, and $16.78$ on Mistral) across multiple conditional reasoning datasets.
We then conduct comprehensive experiments to understand \textit{how} the code representation triggers reasoning abilities and 
\textit{which} capabilities are elicited in the underlying models.
Our analysis on GPT 3.5 reveals that the code formatting of the input problem is essential for performance improvement. Furthermore, the code representation improves \textit{sample efficiency} of in-context learning and facilitates \textit{state tracking} of entities.\footnote{Our code and prompts are available 
\href{https://github.com/UKPLab/arxiv2024-conditional-reasoning-llms}{at this URL.}}

%% file: latex/sections/1-introduction.tex
Reasoning is a fundamental component of both human and artificial intelligence (AI) and the backbone of many NLP tasks. Recently, intensive studies have been performed on different aspects or types of reasoning such as 
mathematical reasoning \citep{patel-etal-2021-nlp, chen-etal-2021-finqa, cobbe2021training}, various kinds of logical reasoning \citep{Liu_2020, logiqa2, sinha-etal-2019-clutrr}, and commonsense-focused reasoning \citep{madaan-etal-2022-language, liu-etal-2022-rainier, liu-etal-2022-generated, wang-etal-2023-elaboration}. \textit{Conditional reasoning}, a primary yet complex reasoning ability that draws alternative conclusions depending on the fulfillment of certain \textit{conditions}, remains understudied. These conditions are stated in the text, making the problem self-contained, which allows us to study the semantic inferencing capabilities of the underlying model, i.e., identifying relevant premises and ascertaining the presence of total evidence \cite{ nolt1988schaum,cabrio2014decomposing} without the requirement for, and confounding effects of external knowledge.
Conditional reasoning is also a fundamental form of reasoning useful in many practical scenarios, such as answering real-world questions about the eligibility for a visa or a loan. Despite the recent introduction of some benchmarks \citep{saeidi-etal-2018-interpretation, sun-etal-2022-conditionalqa, kazemi2023boardgameqa}, conditional reasoning abilities of LLMs remain understudied.

\begin{figure}[t]
\centering
\includegraphics[width=0.47\textwidth]{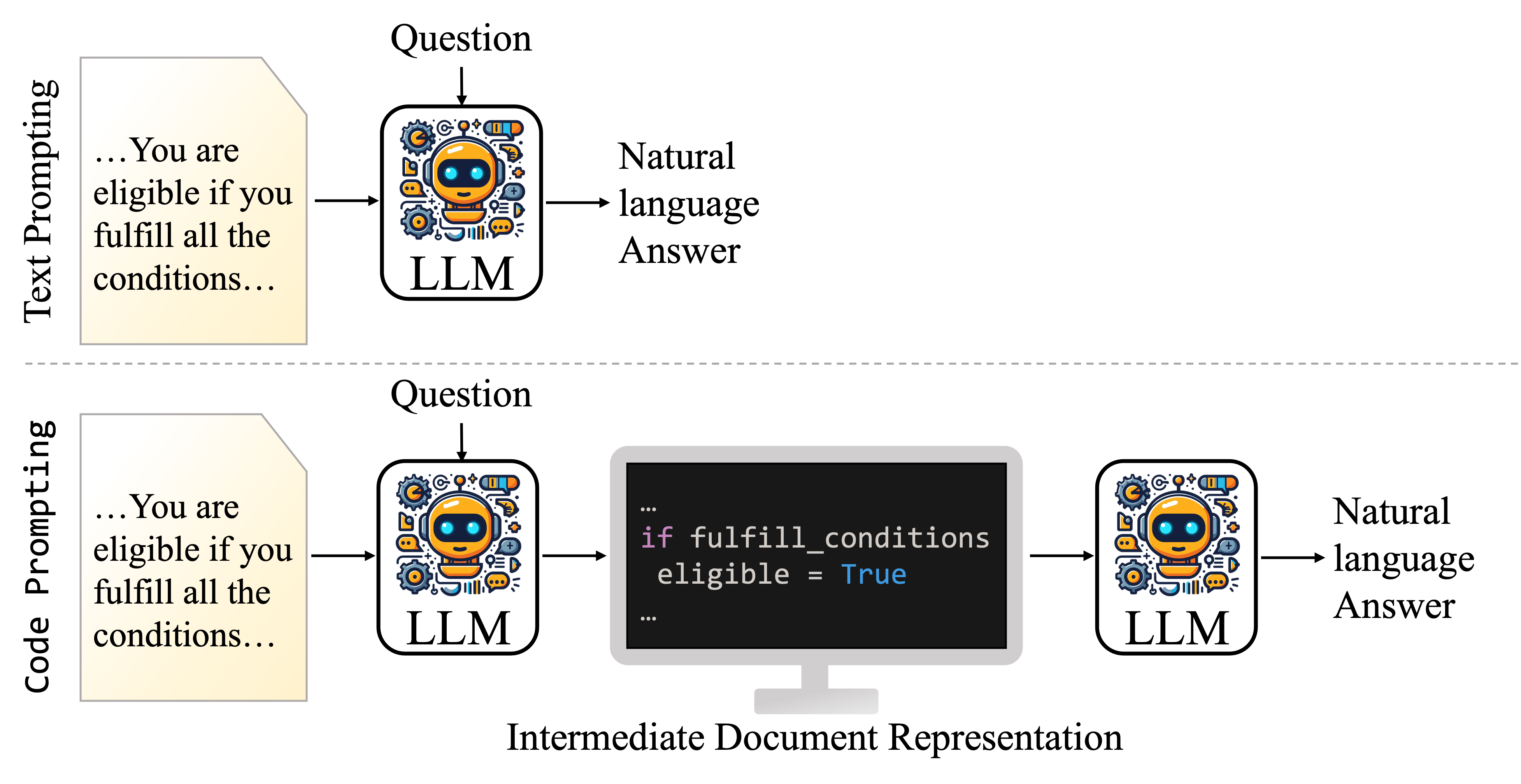}
\caption{Code prompting converts a natural language problem into a \textit{code prompt} and prompts a large language model with such code to generate an answer.}
\label{fig:intro}
\end{figure}

Recently, researchers have analyzed the synergies between LLMs and symbolic interpreters to improve performance on reasoning tasks ~\citep{gao2022pal, chen2022program, lyu2023faithful}. These works transform structured reasoning problems, such as mathematic or symbolic reasoning, into code and run it on an external interpreter. 
In such a setup, LLMs are mainly focused on natural language representation aspects and planning how to solve the problem, while the actual logical reasoning is offloaded to an external execution module, confounding our understanding of the reasoning 
In particular, the fundamental questions of \textit{what} contributes to the reasoning abilities and \textit{how} reasoning abilities are triggered in LLMs remain open. Nevertheless, pretraining on code is considered an important component that contributes to and explains the improved reasoning ability of LLMs. State-of-the-art LLMs such as GPT 3.5 \citep{kojima2022large}, GPT 4 \citep{openai2023gpt4}, Mixtral \citep{jiang2024mixtral}, and Mistral 7B \citep{jiang2023mistral} have been pretrained on both text and code and have demonstrated considerable boosts in many reasoning benchmarks. 

In this work, we analyze whether one can elicit improved conditional reasoning abilities in LLMs by merely changing the input format, i.e., from text to code.
We constrain our experiments to text+code LLMs to run text and code inputs on the same underlying model. In this way, we can avoid the confounding factor of different pretraining data of specialized text and code LLMs.
To understand the benefit of code as an intermediate representation, we devise a chain of prompts, \textit{code prompting}, that transforms a natural language (NL) task into code and directly prompts the LLM with the generated code. The code contains the logical structure needed to solve the problem, along with the original natural language text as code comments. An illustration is provided in \Cref{fig:intro}.
Our contributions are summarized as follows:
\begin{itemize}
    \item We propose a methodology to investigate how the input representation impacts the reasoning abilities of text+code LLMs.
    \item We operationalize such methodology by introducing a \textit{chain of prompts} that transforms a NL task into code, which is then sent back to the LLM to generate NL answers.
    \item We conduct a comprehensive study to compare code prompts with text prompts, showing (i) large performance gains on the three LLMs (up to $22.52$ points for GPT3.5, up to $7.75$ for Mixtral, and up to $16.78$ for Mistral), while (ii) being more efficient with regard to the number of demonstrations.
    \item We conduct extensive analysis to understand why code prompts efficiently elicit conditional reasoning abilities, showing that prompting with code yields largely improved variable state tracking.
\end{itemize}

%% file: latex/sections/2-rw.tex

\paragraph{LLM Types.} We categorize LLMs into three types according to their intended use: i) LLMs for natural language text (\textit{text-only LLMs}), ii) LLMs for coding tasks (\textit{code-only LLMs}), and iii) LLMs for natural language and coding tasks (\textit{text+code LLMs}). The intended use of \textit{text-only LLMs} \citep{zhang2022opt,touvron2023llama} is to process and generate natural language text such as answers to questions.
The intended use of \textit{code-only LLMs} \citep{li2023starcoder, roziere2023code} is to process and generate code.
Lastly, \textit{text+code LLMs} are equally capable of solving natural language and coding tasks. Examples of this are GPT 3.5 \citep{kojima2022large}, Mixtral \citep{jiang2024mixtral}, and Mistral \citep{jiang2023mistral}. In this work, we focus on \textit{text+code LLMs} because of their ability to process two types of input representations interchangeably: natural language text and code. Using such models eliminates the confounding effect of fine-tuning between model variants specialized for only text or code.

\paragraph{Augmenting text with code.} Most works that generate code to solve natural language tasks use an external symbolic interpreter to run the resulting code.
\citet{chen2022program} and \citet{gao2022pal} showed consistent gains on mathematical problems, symbolic reasoning, and algorithmic problems by using LLMs aided by external code interpreters.
\citet{lyu2023faithful} further report improvements in boolean multi-hop QA, planning, and relational inference. In contrast, \citet{Ye-Et-Al:2023:SAT} used an external automated theorem prover with declarative code and showed consistent gains w.r.t. imperative code-interpreter-aided LLMs on arithmetic reasoning, logical reasoning, symbolic reasoning, and regex synthesis tasks. \citet{pan-etal-2023-fact} did not use any interpreter and instead created programs composed of multiple subroutines and used smaller specialized models to run them. In this way, they outperform text prompts on text LLMs for fact-checking tasks. Lastly, \citet{li2023chain} runs pieces of code in an LLM to update the program state when the Python interpreter fails due to a code exception and shows performance gains on BIG-Bench Hard \citep{suzgun2022challenging}.
All these works investigate how to best use an external symbolic interpreter to aid an LLM in solving reasoning tasks, i.e., they \textit{run} code and therefore have a program state with variables and its values. However, we do not employ any external symbolic reasoner, and we \textit{do not run code}. We investigate the reasoning abilities of LLMs under different \textit{input representations} (i.e., text and code). Our code prompts are not executed; they are simply read by the LLM and used to generate a natural language answer.

Some works suggest that code LLMs may possess superior reasoning abilities than text LLMs. \citet{madaan-etal-2022-language} investigate whether code LLMs are superior at \textit{structured} reasoning than text LLMs. They observe that code LLMs can generate graphs that link commonsense concepts better than text LLMs. \citet{liu-etal-2023-magic} investigate code prompts in abductive and counterfactual reasoning tasks and report superior results than text prompts on \texttt{code-davinci} \citep{ouyang2022training}, a code LLM. However, code prompts exhibit mixed results on \texttt{text-davinci-002} \citep{ouyang2022training}, a text LLM. We attribute this to the fact that while this model includes some code in its pretraining corpus, it is not explicitly trained for code generation and, in general, performs poorly on code generation tasks \citep{chen2021evaluating}. Therefore, the effect of the input representation on the reasoning abilities of text+code LLMs remains unclear. Furthermore, the reasons behind the superior performance of code prompts in code LLMs also remain unclear. In our work, we aim to answer whether code prompts can elicit conditional reasoning abilities in text+code LLMs and the reasons behind this.


To the best of our knowledge, only the work of \citet{hussain2023leveraging} investigates the conditional reasoning abilities of LLMs. However, they only analyze the abilities of text LLMs after training them on ConditionalQA \citep{sun-etal-2022-conditionalqa}. 

%% file: latex/sections/3-model.tex
We posit that each LLM encodes a set of capabilities, such as mathematical, logical, or conditional reasoning. However, not all of them are used for every input instance, even if they would be useful. 
We hypothesize that the input representation plays a pivotal role in eliciting such capabilities.
Prior works show that LLMs trained on a combination of text and code exhibit superior reasoning abilities \citep{kojima2022large, openai2023gpt4, jiang2024mixtral, jiang2023mistral}. 
Therefore, we conjecture that a code representation of a natural language (NL) problem may trigger some of these reasoning abilities encoded in text+code LLMs. More formally, we wonder whether exists some space $\mathcal{S}$\footnote{Since the input of LLMs must be strings, $\mathcal{S}$ must be a set of all possible sentences constructed using some alphabet and grammar.} with an associated function $f$ that transforms a natural language problem $p \in \mathcal{N}$ into that space, such that, when prompting an LLM with the representation of $p$ in such space yields better results according to some evaluation function $\sigma$.
\begin{equation*}
    \exists \mathcal{S}, f: \mathcal{N} \rightarrow \mathcal{S},  \sigma(LLM(f(p)) \geq \sigma(LLM(p))
\end{equation*}

We fix $\mathcal{S}$ to the programming language space and define \textit{code prompts} $f(p)$ as prompts that model a natural language problem with code. We also define $f$ as a prompt that transforms the NL text into code.
$f(p)$ code follows the original NL text as much as possible. 
We use a simple structured code that contains the logical structure needed to solve the problem, along with the original NL text as code comments. In particular, it creates variables for key entities in the question and documents and \textit{if blocks} for each conditional statement in the documents. Figure~\ref{fig:code_prompt} exemplifies this transformation and \Cref{appendix:coding_features} provides more details of the code features.
Lastly, we define \textit{code prompting} as $LLM(f(p))$, a chain of prompts that \underline{i}) transform the NL text into code, and \underline{ii}) use this code to prompt the LLM to generate the answer in natural language. \Cref{fig:intro} illustrates this pipeline.

It is important to note that the code is not executed \textit{per se} and therefore, there is no program state. We simply prompt the LLM with the code and ask the LLM to generate a natural language answer based on the content of such code. This setup allows us to investigate the effect of the input representation on text+code LLMs.

%% file: latex/sections/4-setup.tex
\subsection{Task Setup}
We evaluate the conditional reasoning abilities of the LLMs under different prompting methods using the traditional question-answering task. The input is a question and a document, and the model needs to produce the answer, which can be a span of the input document, yes or no. Given that chain-of-thought prompting (CoT, \citealt{wei2022chain}) has been shown to improve the reasoning abilities of LLMs, we instruct the model to generate a CoT before the final answer in all prompting methods. Since we force code prompting to generate a natural language answer, we also force it to generate a natural language CoT.

\subsection{Datasets}
\label{sec:datasets}
Throughout our experiments, we use three question-answering (QA) datasets for conditional reasoning: \textit{ConditionalQA} \citep[\texttt{CondQA};][]{sun-etal-2022-conditionalqa}, a scenario-based question answering (QA) dataset, \textit{BoardgameQA} \citep[\texttt{BGQA};][]{kazemi2023boardgameqa}, a boardgame-base QA dataset with conflicting rules, and \texttt{ShARC} \citep{saeidi-etal-2018-interpretation}, a conversational QA dataset with natural language rules. Solving these datasets requires advanced conditional and compositional reasoning capabilities.

We focus on the QA task of \texttt{CondQA}. For \texttt{BGQA}, we focus on the \textit{main} partition, which includes three subsets \texttt{BGQA-1}, \texttt{BGQA-2}, and \texttt{BGQA-3}, where the number indicates the reasoning hops needed to answer. Lastly, while \texttt{ShARC} encompasses dialogue generation, we aim to evaluate specific capabilities unrelated to conversational flow. Therefore, we isolated the QA pairs from the provided dialogues, resulting in a dataset where the model has to answer \textit{yes}, \textit{no}, or \textit{not enough information}.\footnote{In the full task, \textit{not enough information} would trigger another step in a pipeline to generate a follow-up question.} 
We include more details about the datasets in Appendix \ref{appendix:datasets}, a formal definition of the prompts in \Cref{appendix:prompts}, and examples in \Cref{appendix:prompt_examples}.

\subsection{Models}
\label{sec:models}
We perform our study using text+code LLMs because of their ability to process text and code interchangeably. We do not employ code-only LLMs because their intended use does not include solving natural language tasks \citep{roziere2023code}, as required in our case. Similarly, we do not employ text-only LLMs because they cannot generate code. Furthermore, using text+code LLMs also allow us to eliminate the confouding effect of fine-tuning between model variants specialized for only text or code. We corroborate the shortcomings of text-only and code-only LLMs with additional experiments on CodeLLaMA \citep{roziere2023code} and LLaMA 2 \citep{touvron2023llama2openfoundation} in \Cref{appendix:codellm} and \ref{appendix:textllm}.

We employ OpenAI's \texttt{gpt-35-turbo}, \texttt{Mixtral 8x7B} \citep{jiang2024mixtral}, and \texttt{Mistral 7B} \citep{jiang2023mistral}. The use of these models allows us to investigate whether our hypothesis holds across all available sizes of text+code LLMs. 
We execute our prompts with in-context learning and provide one demonstration per class. More details on the LLM setup are provided in \Cref{appendix:llm_setup}. 

\subsection{Evaluation} We follow the evaluation metrics used in the original datasets. For \texttt{CondQA}, we report the F1 token overlap between the predicted answer and the label, while for \texttt{BGQA} and \texttt{ShARC}, we report the macro F1 score. 
We run the main experiments two times with different random seeds ($0$ and $1$). We report the average and standard deviation performance across these runs. For the subsequent analyses of code prompts, we run each experiment once only on GPT 3.5 due to the inference costs.

%% file: latex/new_tables/overall_results.tex
\begin{table*}[t]
\renewcommand{\arraystretch}{0.8}
\begin{center}
\scalebox{0.88}{
\begin{tabular}{llcccccccc}
\toprule
\textbf{Model} & \textbf{Prompt}                   & \textbf{CondQA} & \textbf{ShARC}     & \textbf{BGQA-1} & \textbf{BGQA-2} & \textbf{BGQA-3} & $\Delta$\textbf{CP} \\
\midrule
\multicolumn{8}{c}{\textbf{Test Set}} \\
\midrule
\multirow{2}{*}{GPT 3.5} &Text  & $58.70$ & $\mathbf{62.95}$ & $51.15$  & $37.42$  & $27.77$ & \multirow{2}{*}{8.42} \\
                         &Code  & $\mathbf{60.60}$ & $54.98$ & $\mathbf{58.67}$ & $\mathbf{55.56}$ & $\mathbf{50.29}$\\

\midrule

\multirow{2}{*}{Mixtral} &Text & $\mathbf{48.17}$ & $53.77$ & $\mathbf{56.38}$ &$ 39.64$  &$ 30.15$ & \multirow{2}{*}{4.22}\\
                         &Code & $44.73$ & $\mathbf{59.06}$ & $53.33$ & $\mathbf{47.39} $&  $\mathbf{44.72}$ \\

\midrule

\multirow{2}{*}{Mistral} &Text & $\mathbf{35.74}$  & $43.60$ & $47.40 $ & $48.78$  & $47.86$ & \multirow{2}{*}{2.74} \\
                         &Code & $33.28$ & $\mathbf{49.92}$  & $\mathbf{53.80}$ & $\mathbf{51.27}$  & $\mathbf{48.79}$ \\

\midrule
\multicolumn{8}{c}{\textbf{Dev Set}} \\
\midrule
\multirow{2}{*}{GPT 3.5} &Text           & $56.54\pm 0.08 $ & $\mathbf{64.10} \pm \mathbf{0.10}$ & $53.16 \pm 1.67$ & $33.71 \pm 10.37$  & $31.5 \pm 13.39$ & \multirow{2}{*}{9.84} \\
&Code           & $\mathbf{57.64 \pm 1.42}$ & $58.54 \pm 1.22$ & $\mathbf{68.60 \pm 1.09}$ & $\mathbf{55.85 \pm 4.06}$ & $\mathbf{47.57 \pm 2.68}$\\
\midrule

\multirow{2}{*}{Mixtral} &Text & $\mathbf{46.60} \pm \mathbf{0.99}$ & $55.71 \pm 2.51$ & $58.31 \pm 1.77$  & $36.77 \pm 0.09$ & $32.06 \pm 1.79$  & \multirow{2}{*}{2.51}\\
&Code & $40.88 \pm 1.84$ & $\mathbf{58.96} \pm \mathbf{3.44}$ & $57.94 \pm 5.52$ & $\mathbf{45.32} \pm \mathbf{0.54}$ & $\mathbf{38.90} \pm \mathbf{7.33}$  \\
\midrule

\multirow{2}{*}{Mistral} &Text & $28.84 \pm 0.02$ & $37.56 \pm 0.78$ & $47.61 \pm 0.92$ & $47.29 \pm 1.97$  & $46.56 \pm 2.92$ & \multirow{2}{*}{5.10}\\
&Code           & $28.26 \pm 10.03$ & $\mathbf{53.42} \pm \mathbf{0.93}$  & $\mathbf{52.21} \pm \mathbf{0.95}$ & $\mathbf{54.27} \pm \mathbf{1.42}$ & $45.22 \pm 10.75$  \\

\bottomrule
\end{tabular}
}
\end{center}
\caption{Comparison (F1 score) of text prompt and code prompts. All results use one demonstration per class. ~~~$\Delta$CP = Code Prompt - Text Prompt, i.e.,  the average performance gain from code prompts across all datasets.}
\label{table:main_results}
\end{table*}

%% file: latex/sections/5-experiments.tex
\label{sec:experiments}
We devise a set of experiments to analyze and quantify whether the code representation of a natural language prompt (i.e., code prompts) elicits conditional reasoning abilities and why.
We first compare the performance of the two prompting methods --- \textit{text prompts} and \textit{code prompts} on three LLMs across three datasets ($\S$\ref{sec:codevstext}).
We then conduct extensive ablation experiments on the dev set of the datasets with GPT 3.5, the best-performing and largest model, to understand the reason behind the performance gain from code prompting.
In particular, we study whether \textit{code syntax} or the implicit \textit{text simplification} from the code translation is what improves performance (\Cref{sec:code-imp}).
We also check if the improvement is caused by the models merely being exposed to code within prompts and not necessarily the instances translated to code (\Cref{sec:ablation}).
Furthermore, we show that code prompting is more \textit{sample efficient} (\Cref{sec:efficiency}) when compared to text prompting and that models prompted with code exhibit superior \textit{state tracking} capabilities (\Cref{sec:variable-tracking}).
Lastly, we conduct a human evaluation that confirms the faithfulness of the generated code in \Cref{appendix:human_eval}.

\subsection{Code Prompting Improves over Text Prompting}
\label{sec:codevstext}
\input{latex/sections/5-1-overall_results}

\subsection{Code Syntax Elicits Reasoning Abilities}
\label{sec:code-imp}
\input{latex/sections/5-2-code_format}
\input{latex/new_tables/code_format}

\subsection{Code Semantics are Important}
\label{sec:ablation}
\input{latex/sections/5-3-code-semantics}
\input{latex/tables/code_semantics}

\subsection{Code Prompts are More Sample-Efficient at Eliciting Reasoning Abilities}
\label{sec:efficiency}
\input{latex/sections/5-4-sample_efficiency}

\subsection{Code Prompts Improve Variable Tracking in LLMs}
\label{sec:variable-tracking}
\input{latex/sections/5-5-variable_tracking}


%% file: latex/sections/5-1-overall_results.tex
\Cref{table:main_results} shows the model performance on the development and test sets. Code prompts outperform text prompts in the majority of cases on the test set (11 out of 15). This trend holds true across models, with each achieving peak performance through code prompts for most datasets (i.e., GPT-3.5 in 4/5, Mixtral in 3/5, Mistral in 4/5). Notably, code prompts consistently surpass text prompts on \texttt{BGQA-2} and \texttt{BGQA-3}, the most reasoning-intensive datasets (see \Cref{appendix:datasets}), for all models. This is particularly evident for GPT-3.5, where gains exceed 18 points. Conversely, the advantage is narrower on \texttt{CondQA}, where the linguistic dimension plays the biggest role (see \Cref{appendix:datasets}). This suggests that code prompts elicit conditional reasoning abilities and are most suited for reasoning-intensive tasks. Furthermore, in the cases where text prompts are superior, their average gains are only 4.23. In contrast, code prompts lead to significantly larger mean gains of 8.53 in the cases where they are superior. Additionally, an experiment with Phi-2, a small language model, reveals a substantial 15-point performance improvement using code prompts (see Appendix \ref{appendix:phi2}). 

To shed light on the performance gains driven by code prompts, we delve into the confusion matrices (attached in \Cref{appendix:confusion_matrix}) and discover that text prompts in Mistral predict ``not enough information'' much less than code prompts for \texttt{BGQA}. This is particularly noticeable in \texttt{BGQA-1}, where text prompts do not predict a single ``not enough information,'' while code prompts do. On the other hand, text prompts in GPT 3.5 and Mixtral overpredict ``not enough information'' on \texttt{BGQA} and \texttt{ShARC}, leading to a low number of true positives for the conclusive answers. We hypothesize that this model hesitation could stem from the \textit{alignment tax} \citep{ouyang2022training} of \textit{reinforcement learning from human feedback} models. This potential barrier may be alleviated by code prompts because they indicate to the model the variable that answers the question and instruct the model to track the entailment status of variables within the given code.

These consistent and substantial gains from code prompts are obtained despite a straightforward transformation of text prompts, which does not incorporate new information, as shown in \Cref{fig:code_prompt}. 
This finding strongly suggests that code possesses specific characteristics that effectively elicit conditional reasoning abilities within text+code LLMs.

%% file: latex/sections/5-2-code_format.tex
We now want to delve into the source of the performance gains observed when using code prompting. We investigate whether these improvements stem from the simplification of text into premises facilitated by code, effectively reducing the task to a form of semantic inference within the \textit{linguistic dimension}, or if there are inherent properties of code syntax that contribute to enhanced performance. To investigate this, we devise experiments with prompts that represent the intermediate states between natural language and code.

\noindent
I. \textbf{Atomic Statements.}~~~Inspired by \citet{min-etal-2023-factscore}, we transform each NL sentence\footnote{We only transform the \textit{facts} in \texttt{BGQA} since transforming the \textit{rules} into atomic statements as well yields worse results.} into a sequence of \textit{atomic statements}, which we then append to the original sentence. In this way, the atomic statements can be seen as defining variables for each key entity in the text. Hence, this new prompt would resemble code but without control flow and in natural language form. 
The prompt retains access to the original instance text (i.e., no loss of information) but is also augmented by simplified sentences in the form of atomic statements. 
This setup allows us to investigate whether the \textit{simplicity} of the input triggers improves reasoning abilities, regardless of the text and code syntax. 

\noindent
II. \textbf{Back-Translated Code.}~~~In our second experiment, we investigate whether the \textit{semantics} of the code statements and not the code \textit{syntax} are the reason behind the performance boost. 
For this purpose, we back-transform the code prompts into NL such that the reasoning statements (i.e., the \textit{if} conditions) are clearly and concisely stated in natural language.
Specifically, we map every variable into the format \textit{Key entity: variable without snake case.} For instance, the variable \textit{husband\_pass\_away} from \Cref{fig:code_prompt} would be back-transformed as \textit{Key entity: husband pass away.}
To transform the \textit{if} statements, we create a translation prompt by providing four demonstrations. 
These demonstrations simply translate the conditional statements within the ~code-formatted instance back into natural language. We also translate the variables in the same manner. This makes the back-translated text as close as possible to the code text. We provide examples of this in \Cref{table:back_translation_example} from \Cref{appendix:code_ablations}.

\noindent
\textbf{Results.}~~~The results\footnote{We do not conduct ablation tests on ShARC because, as explained in \Cref{sec:experiments}, these ablations aim to understand why code prompts outperform text prompts using the highest performing model. Results for Mistral and Mixtral are shown in \Cref{appendix:ablations_local_llms}.} 
in \Cref{table:code_format} show that (\underline{1}) prompting with atomic statements does not reach the performance of code prompts, and (\underline{2}) mapping back from code to NL results in a performance drop compared to code prompts. 
These findings suggest that code prompts enhance LLM performance beyond mere text simplification. This conclusion is supported by the observation that these alternative text simplification approaches, despite offering similar semantics to code prompts, fail to replicate the performance gains observed with code prompts. Therefore, these results imply that specific syntactic features embedded within code directly contribute to performance improvement.

Lastly, our evaluation on \texttt{BGQA-3} reveals a significantly larger performance decline when using atomic statements compared to back-translated code. This disparity likely stems from the dataset's inherent structure. The method we employ for generating atomic statements \citep{min-etal-2023-factscore} was specifically designed for general text formats like Wikipedia pages. However, \texttt{BGQA} is a logic-based dataset where input "facts" are already presented as minimally informative statements, deviating from the typical structure of general documents. As a result, generating atomic statements from these sentences can unintentionally disrupt the sentence structure, making it difficult to track the attributes of subjects and objects within the text. This observation is further supported by our results on \texttt{CondQA}, a dataset with longer documents, where atomic statements achieve higher performance than back-translated code.

%% file: latex/new_tables/code_format.tex
\begin{table}[t]
\centering
\renewcommand{\arraystretch}{0.85}
\setlength{\tabcolsep}{3pt}
\begin{tabular}{l c c}
\toprule
\textbf{Dataset}  & \textbf{$\Delta$ Atomic St.} & \textbf{$\Delta$ Code $\rightarrow$ NL} \\
\midrule
CondQA & $-2.66$ & $-4.72$ \\
BGQA-1 & $-4.37$ & $-1.43$ \\
BGQA-2 & $-8.72$ & $-5.39$ \\
BGQA-3 & $-19.26$ & $-3.68$ \\
\bottomrule
\end{tabular}
\caption{Performance gap of \textit{atomic statements} and \textit{back-translated code} when compared to code prompts using GPT 3.5. Results from the dev set of each dataset.}
\label{table:code_format}
\end{table}


%% file: latex/sections/5-3-code-semantics.tex
Previously, we have shown that code syntax is necessary to elicit the reasoning abilities of text+code LLMs. Now, we aim to investigate which aspects of code are pivotal. In particular, we evaluate the impact of retaining the natural language text of the original instance within the code comments and the importance of the code semantics.
To analyze the former, we have (\underline{1}) removed the code comments that include the original natural language text from the input and evaluated the performance of the new prompts. To analyze the latter, we (\underline{2}) perturbed the code to anonymize the variables and functions, as well as (\underline{3}) added random code whose semantics are completely irrelevant to the original natural language text. In the latter two cases, the code comments remain unmodified (examples illustrating them are provided in \Cref{table:code_ablations_example} from \Cref{appendix:code_ablations}).
Since \texttt{CondQA} includes span answers and removing the NL text would make it impossible for the model to generate the span, we only report performance on the yes-no answers partition (\texttt{CondQA-YN}). 

Table \ref{table:code_semantics} shows that removing the NL text in the code comments yields a performance drop of $14.02$ points on \texttt{CondQA} and a performance drop between $16.7$ and $5.2$ on \texttt{BGQA}. This significant and consistent decrease in all datasets confirms that retaining NL text in comments is vital for the LLM to understand the input problem.

\paragraph{Effect of Code Perturbations.} Code perturbations (\textit{anonymous code} and \textit{random code}) confirm the importance of code semantics in eliciting reasoning abilities. When we use anonymized code, we observe a performance reduction of almost $2$ points on \texttt{CondQA} and a decrease between $6.6$ and $4$ in \texttt{BGQA}. The decrease is even larger when the code is randomized, with drops of more than $3$ points on \texttt{CondQA} and between $7.4$ and $9.8$ on \texttt{BGQA}. This more pronounced drop is expected since the semantics and logic of the code mismatch the NL text, whereas anonymous code maintains the same logic on both NL and code. 
Furthermore, we also observe that the performance drop of random code prompts is similar to that of text prompts (Table~\ref{table:main_results}) on \texttt{CondQA} and \texttt{BGQA-1}. This can be interpreted as the model being able to identify the irrelevance of the code to the text. Hence, the model disregards the code to solely focus on the code comments (i.e., the natural language text). This could be possible thanks to the provided demonstrations, which show answers that only refer to the natural language text.

These results confirm that code alone does not trigger reasoning abilities, and instead, the combination of code that represents the original natural language instance and the NL text is able to unlock the potential of LLMs. We observe similar patterns on Mistral and Mixtral in \Cref{appendix:ablations_local_llms}.

%% file: latex/tables/code_semantics.tex
\begin{table}[t]
\begin{center}
\renewcommand{\arraystretch}{0.85}
\setlength{\tabcolsep}{2.3pt}
\scalebox{0.85}{
\begin{tabular}{lccccccc}
\toprule
\textbf{Prompt}                   & \textbf{CQA} & \textbf{CQA-YN}     & \textbf{BG\textsubscript{1}} & \textbf{BG\textsubscript{2}} & \textbf{BG\textsubscript{3}} \\
\midrule
Anonym. & $-1.62$ & $-2.90$ & $-6.60$ & $-4.80$ & $-4.00$ &\\ 
Random  & $-3.40$ & $-2.67$ & $-7.40$ & $-9.20$  & $-9.80$ &\\				
- Comments & N.A. & $-14.02$ & $-16.70$ & $-16.20$ & $-5.20$ \\			

\bottomrule
\end{tabular}
}
\end{center}
\caption{Performance gap to code prompts for each code perturbation. cQA stands for CondQA, CQA-YN for the partition of CondQA with yes-no answers, BG for BGQA. Results reported on the dev set of each dataset.}
\label{table:code_semantics}
\end{table}



%% file: latex/sections/5-4-sample_efficiency.tex
Given our observations that code prompts trigger conditional reasoning abilities better than text prompts, it is natural to ask the follow-up question: are code prompts also more \textit{sample-efficient} than text prompts? To answer this, we evaluate how the overall performance of GPT 3.5 changes with respect to the number of demonstrations for the two prompting methods.

\begin{figure}[h]
\centering
\includegraphics[width=0.45\textwidth]{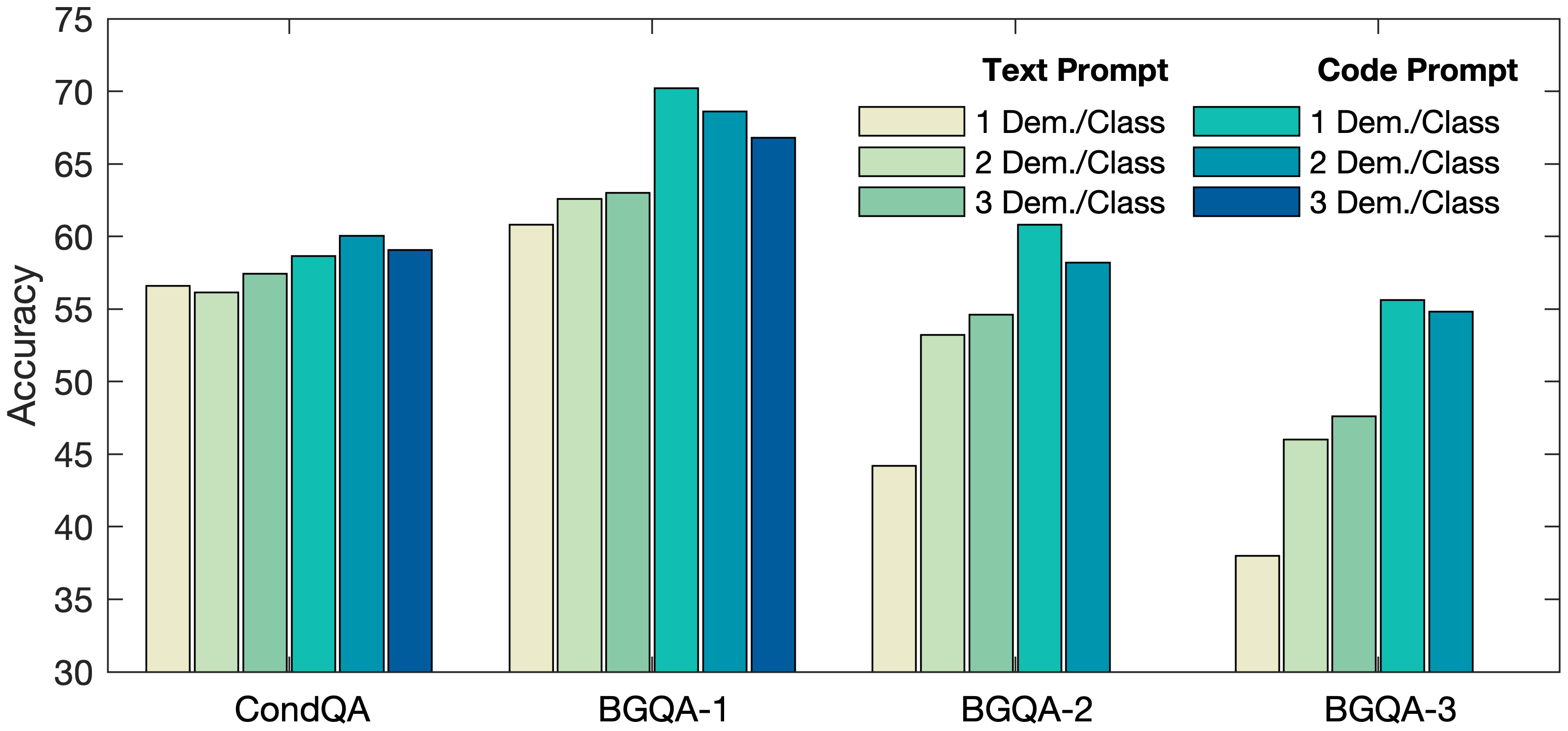}
\caption{Performance comparison of GPT 3.5 between text (green) and code prompts (blue) using $1$, $2$, and $3$ demonstrations per class. Results reported on dev sets.}
\label{fig:few_shot}
\end{figure}

\Cref{fig:few_shot} shows that when we only provide one demonstration per class (i.e., answer type in our datasets), the performance gap is the largest across all datasets. As expected, this gap decreases when we provide more demonstrations. Moreover, we also observe that code prompts with only one demonstration per class even outperform text prompts with three demonstrations per class, which further shows the sample efficiency of code prompts.
These results indicate that code prompts trigger conditional reasoning more efficiently than text prompts on GPT 3.5, and this is one of the reasons for its superior performance. We conduct additional analysis on Mistral and Mixtral in \Cref{appendix:demonstrations_local_llms}.

%% file: latex/sections/5-5-variable_tracking.tex
We hypothesize that one of the reasons for the superior performance of code prompting is an improved ability to identify and track the states of key variables or concepts. 
This hypothesis is based on the intuition that, for natural language in general, local context is the most important part to generate the next token \cite{khandelwal-etal-2018-sharp, sun-etal-2021-long}. However, generating code is often more challenging because code frequently refers to previously defined functions and variables, which can be dozens or even hundreds of lines apart. This resembles multi-hop reasoning, where the model may need to reference a key entity dozens of lines before. Therefore, an improved ability to \textit{look for distant co-references} caused by training on code can be beneficial for multi-hop reasoning, which is also needed to solve our datasets.

To test our hypothesis, we devise the following experiment.
Firstly, we define \textit{reasoning step} as each output sentence split by ``\textbackslash n.'' After generating each reasoning step, we \textit{stop} the model generation and query about all key entities defined in the input prompt. In the case of text prompts, we query the model whether the given facts are true or not, and for code prompts, we query for the value of the (boolean) variables. 
In all cases, the model only has to generate \textit{True}, \textit{False}, a \textit{string}, or \textit{unknown}. Then, we compare the percentage of errors in text and code prompts. This number represents the \textit{memory errors} committed by the model. The more memory errors there are, the more difficult it is for the model to track and remember entities/variables.
We provide further details on how we extracted the key entities to ask for, how we identified the reasoning steps in the chain of thought used to stop the model for conducting the probes, and examples of the prompt probes in \Cref{appendix:variable_tracking} and its \Cref{table:variable_tracking_example}.

\input{latex/tables/variable_tracking_v2}

\noindent
\textbf{Does Generated Text reflect Model Beliefs?}~~~As the generated text may not be faithful to the internal beliefs of the model \citep{lyu2023faithful}, we first test the validity of this experiment as a proxy metric of the internal belief of the model. 
To do this, we compare the memory error percentage of the prompting methods in instances where the model solves (i.e., \textit{correct instances}) and does not solve (i.e., \textit{incorrect instances}) the question.
If incorrect instances yield a higher memory error, this would indicate that the model struggles more to remember the variable states on those instances, which in turn would make it more likely to fail when conducting the reasoning process. Therefore, our probes would be a proxy metric of the internal belief of the model.

\Cref{tab:variable_tracking} shows the results of this comparison. We observe that all prompting methods in all datasets consistently make more memory mistakes on incorrect instances than on correct instances, with the exception of text prompts on \texttt{CondQA}.
However, the memory error in this case is significantly high, which may suggest that the model is not able to track entities correctly in either case. 
Therefore, we can use this experiment as a proxy measure of the memory of the model.

\noindent
\textbf{Code Prompting improves State Tracking.}~~~From \Cref{tab:variable_tracking}, we further observe that Text Prompts make significantly more memory errors than code prompts on all datasets on GPT 3.5 (Results for Mistral and Mixtral are provided in \Cref{appendix:memory_errors_local_llms}). Specifically, the gap is consistently more than 30\% with peaks on \texttt{CondQA} (66.69\%) and \texttt{BGQA-3} (39.8\%). Therefore, this experiment empirically confirms our hypothesis that code prompts improve state tracking of the key entities and variables when compared to text prompts.

%% file: latex/tables/variable_tracking_v2.tex
\begin{table}[t]
    \centering\renewcommand{\arraystretch}{0.85}
    \begin{tabular}{ccc|cc}
    \toprule
         &  \multicolumn{2}{c}{\textbf{Correct Ans.}}&  \multicolumn{2}{c}{\textbf{Incorrect Ans.}}\\
         \textbf{Dataset}&  \textbf{Text}&  \textbf{Code}&  \textbf{Text}& \textbf{Code}\\
         \midrule
         CondQA  & 71.08 & \textbf{4.39} & 60.79 & \textbf{11.39}\\
         BGQA-1 & 39.33  & \textbf{8.84} & 51.65 & \textbf{22.12}\\
         BGQA-2 & 44.79  & \textbf{15.04} & 52.54 & \textbf{24.75}\\
         BGQA-3 & 54.01 & \textbf{14.21} & 52.13 & \textbf{16.98}\\
    \bottomrule              

    \end{tabular}
    \caption{Comparison of the percentage of memory errors made by GPT 3.5. For each dataset, we separately compute memory errors for the instances where the model gives the correct and incorrect answers. Lower is better. Results from the dev set of each dataset.}
    \label{tab:variable_tracking}
\end{table}

%% file: latex/sections/appendix/human_evaluation.tex
We conduct a small human evaluation to confirm the faithfulness of the generated code to the source natural language text. We evaluate the code generations of all our models on ten random samples from the dev set of \texttt{CondQA}, \texttt{ShARC}, and \texttt{BGQA} (in particular, we use \texttt{BGQA-1} partition). We check for perfect translations, and for the failing cases, we analyze the errors.

\paragraph{\texttt{BGQA.}} We observe perfect translations in all models for all the analyzed samples. We attribute this effectiveness to the close alignment between the natural language documents and first-order logic.

\paragraph{\texttt{ShARC.}} We observe that \texttt{GPT 3.5} generates perfect translations in all cases except one. However, this case is a corner case where the document is irrelevant to the question, and therefore, there is no answer. Furthermore, the document is only one line. Consequently, the model does not generate code and simply keeps the text as a code comment. In the case of \texttt{Mixtral 8x7B}, we observed perfect code translations for 70\% of the samples. One of the failing cases assings as the question variable a variable that is actually from the conversation history, no the quesiton. Another error case exhibits wrong value assingments to some variables. They should be \texttt{none}, but they are assinged \texttt{true} and \texttt{false}. The last case is the corner case explain above. As for \texttt{Mistral 7B}, we find that 60\% of the analyzed samples have a perfect translation. In the remaining 40\%, we observe three cases with no question variable and the same corner case as before. However, the semantics of the natural language text remain, thanks to the code comments.

\paragraph{\texttt{CondQA.}} We observe that \texttt{GPT 3.5} generates a perfect translation in 8 out of the 10 cases. In these two cases with errors, we observe that most of the code is correct, but in both cases, one conditional statement is missed. The model directly generates the body of the if statement without the corresponding if. It is also worth noting that the code is of high quality, including data structures as ditionaries, generating code that explains tables, and also generates lists of strings. In the case of \texttt{Mixtral 8x7B}, we obtain perfect translations in 6 out of 10 cases. All the failing cases exhibit the same type of error: there is a variable statement without a prior if condition. It is worth noting that the code, in general, is of high quality, contains data structures such as dictionaries and even process tables. Lastly, in the case of \texttt{Mistral 7B}, we observed a bit worse results. Only 4 out of the 10 cases are perfect translations. In two cases, there is no code, and instead, the model only generated the original natural language text in code comments. We also observe one case where the first half of the text is correctly translated into code but the second half only contains the code comments representing the natural language text. We also observe one case where the code is correct, but the indentation is wrong; all code blocks are under the first if statement, which should not be like that. Lastly, we find two cases where the if statements do not contain an execution body. Notably,  even in the cases where the code is not perfect, the original semantics from the natural language remain untouched because they are preserved through code comments.

This analysis 
confirms that,
in general, the translated code faithfully represents the semantics of the source natural language text.

%% file: latex/sections/6-conclusions.tex
This work demonstrates that the code representation of a natural language task (i.e., code prompts) can elicit reasoning abilities in large language models (LLMs) of text and code.
These code prompts contain the original natural language (NL) formulation as a code comment and code that formulates the logic of the text. To create these code prompts, we use in-context learning to teach an LLM how to conduct such a transformation automatically. Through multiple experiments on three LLMs and three datasets, we show that code prompts trigger conditional reasoning abilities, with large performance gains w.r.t. text prompts (up to $22.52$ percentage points on GPT 3.5, $7.75$ on Mixtral, and $16.78$ Mistral).
Our experiments reveal that even simple code can be beneficial as long as it closely follows the semantics of the NL text and is accompanied by the original NL text. We also show that code prompts are more sample-efficient than text prompts and that their performance boost stems from their superior ability to identify and track the state of key variables or entities, a central aspect of the logical dimension of semantic inference.

In our future work, we plan to extend to a wider range of reasoning abilities, such as multi-hop reasoning, to understand the capacity and generalizability of code prompting. We also plan to investigate how pretraining on text, code, and text+code affects the triggering of LLMs' reasoning abilities.

%% file: latex/sections/limitations.tex
Transforming a natural language problem into code requires an intermediate step that raises the cost of the whole pipeline. However, this mapping is not a complicated task, as even the smallest models we considered were able to perform it successfully in an in-context learning setup. Therefore, we believe it would be possible to train a small generative model to do it instead of using a large language model. In this way, we could minimize the cost of using code prompts without affecting its performance.

We only ran the experiments on the dev set with two different random seeds due to the costs of running large language models and because we prioritized experimenting on multiple models and datasets. Nevertheless, the results of all models exhibit similar patterns, which confirms the representativeness of our results.
Also, we conduct the ablations only on GPT 3.5, the best-performing and largest model. However, confirming that the findings from these ablations also hold on the smaller models would be interesting.

This work focuses only on analyzing the effects of code representations for natural language tasks. However, it could be possible that other input representation spaces also elicit reasoning abilities. We limit the scope of this work to only the space of simple structured languages (more details on \Cref{appendix:coding_features}) because prior research suggests that pretraining on code improves the reasoning abilities of LLMs, but it might be possible that certain natural languages such as German or Chinese, or certain types of programming languages, such as declarative or logical, also elicit certain abilities. Similarly, we do not conduct experiments on multiple code generation methods because our goal is to analyze whether the mere change of the representation can elicit reasoning abilities and not an analysis of the best coding style.

Our tasks require instruction following abilities, so we do not conduct comparisons of base vs. chat models. Future work could investigate whether instruction tuning has an impact on the LLMs' abilities to understand code.


Lastly, we conduct our experiments on data in English. Analyzing whether our findings hold true in other languages would be interesting. However, the lack of conditional reasoning datasets in other languages would make this study difficult.

%% file: latex/sections/ethics.tex
Our work aims to improve the reasoning abilities of LLMs. The use of code prompts may also simplify the explainability of the model responses since we can inspect the entailment status of the variables. We hope these results contribute to enhancing the trustworthiness and safety of LLMs. Nevertheless, every development may pose some risks. In our case, the improvement of the reasoning abilities in LLMs may be utilized by malicious actors to propagate more persuasive disinformation.

%% file: latex/sections/appendix/datasets.tex
\paragraph{\texttt{ConditionalQA}} is a QA dataset where the answers are applicable under specific scenarios (i.e., conditional answers). Therefore, along with each question, the dataset provides a scenario that describes the background of the person posing such a question. Questions require multi-hop, compositional, and conditional logic over documents about public policies (e.g., the eligibility for a subsidy). Answers can be a span of the document, \textit{yes}, and \textit{no}. We use an oracle retriever to select the relevant passages to the question so that we can isolate the analysis of conditional reasoning abilities in LLMs from the retrieval component. 
The expected output is a chain of thought (CoT; \citealt{wei2022chain}) followed by the final answer. To create the CoT, we use the annotated evidence sentences. We use an oracle retriever to retrieve the relevant passages to the question. This retriever is based on the sentences annotated as evidence for the answer (i.e., rationales). We concatenate all sections that include one rationale and use the resulting passage as input document.

\paragraph{\texttt{BoardgameQA}} is a dataset that evaluates the ability to reason with contradictory information guided by preferences. For example, given a question about traveling abroad, information found online about regulations can be contradictory because rules may change over time. Answering questions in this dataset requires complex multi-hop reasoning with conditional, deductive, and compositional abilities. The domain of the problems is board games, which allows us to analyze the conditional reasoning abilities in a completely different domain from \texttt{CondQA}. \texttt{BGQA} is divided into multiple partitions focusing on different characteristics, such as the depth of the reasoning tree, the need for external information, etc. We focus on the \textit{main} partition and its subpartitions (i.e., \texttt{BGQA-1}, \texttt{BGQA-2}, \texttt{BGQA-3}), where the number refers to the number of reasoning hops required to answer the question. 
This dataset also includes annotated chain-of-thoughts (CoT); therefore, we use their annotated input (``\textit{example}'') as the input prompt and their annotated CoT (``\textit{proof}'') as the expected output.

\paragraph{\texttt{ShARC}} is a conversational QA dataset with natural language rules where most questions are underspecified. Therefore, the model may need to ask a follow-up question to know more about the background of the interlocutor to return an answer. The documents are of legal domain retrieved from the web pages of different governments and state agencies. Since this is a conversational QA and we are not interested in evaluating the conversational abilities of LLMs, we transform the task into regular QA, instead of conversational QA. To do this, the model must answer \textit{yes}, \textit{no}, or \textit{not enough information} for each question. In the original task, \textit{not enough information}, would lead to the generation of a follow-up question.

\paragraph{Complexity of the datasets.} We analyze the complexity of the datasets by counting the percentage of reasoning operations (i.e., \textit{if statements}) in the code prompt generated by GPT 3.5. This analysis shows that the most difficult dataset is \texttt{BGQA-3} with $21.58$\% of reasoning operations, followed by \texttt{BGQA-2} ($16.99$\%), \texttt{CondQA} ($14.66$\%), \texttt{BGQA-1} (\texttt{$10.55$\%}), and lastly, \texttt{ShARC} ($8.32$\%).

We also analyze the length of the documents of each dataset and find that \texttt{BGQA-3} has the longest documents with an average of $39$ lines of code, followed by \texttt{CondQA} ($38$), \texttt{BGQA-2} ($25$), \texttt{ShARC} ($22$), and lastly \texttt{BGQA-1} ($15$). It is worth noting that the documents from \texttt{CondQA} are the short documents extracted with the oracle retriever described above, instead of the full documents, which are much longer (up to $9$k tokens).

These two analyses suggest that \texttt{BGQA-3} and \texttt{BGQA-2} are the most reasoning-intensive datasets due to the high proportion of reasoning operations. In contrast, \texttt{CondQA} is the dataset where the linguistic dimension plays the biggest role because their documents are among the longest ones while it contains much less proportion of reasoning operations than the other datasets with similar document lengths.

\paragraph{Dataset sizes, licenses, and safety.}The sizes and licenses of all the datasets used in this work are provided in \Cref{table:dataset_sizes}. Our use of these datasets is consistent with their intended use, i.e., academic research to evaluate question-answering systems. As far as we know, these datasets do not contain any personal information or offensive content. Although we did not explicitly analyze this, the authors of these datasets did not mention including such content, and we did not observe such content during our use of the datasets. All these datasets are in English.

\begin{table}[h]
    \centering
    \renewcommand{\arraystretch}{0.85}
    \setlength{\tabcolsep}{2.7pt}
    \begin{tabular}{ccccc}
    \toprule
        \textbf{Dataset} & \textbf{Training}  & \textbf{Dev} & \textbf{Test} & \textbf{License} \\
        \midrule
         CondQA & 2338 & 285 & 804 & BSD 2\\
         BGQA-1 & 1000 & 500 & 1000 & CC BY\\
         BGQA-2 & 1000 & 500 & 1000 & CC BY\\
         BGQA-3 & 1000 & 500 & 1000 & CC BY\\
         ShARC & 21890 & 2270 & 8276 & CC-BY-SA-3.0\\
        \bottomrule
    \end{tabular}
    \caption{Sizes of the datasets.}
    \label{table:dataset_sizes}
\end{table}

%% file: latex/sections/appendix/prompts.tex
\paragraph{\texttt{CONDQA}.} Firstly, we define the different components of a data point: scenario ($S$), question ($Q$), document ($D$), rationales ($R$), and answer ($A$). Then, the text prompt $tp$ is defined as follows:
\begin{equation}
\begin{split}
tp = \text{"Question:"} + S + Q + \text{"Document:"} + D \\+ \text{"Let's think step by step"}
\end{split}
\end{equation}

where $+$ represents the string concatenation operator. Then, the output format, $to$ is:
\begin{equation}
    to = R + \text{"Answer:"} + A
\end{equation}

For code prompts, we first define a function ${C: \mathbb{NL} \rightarrow \mathbb{C}}$ that maps a natural language text into code as shown in \Cref{fig:code_prompt}. Then, we define code prompt $cp$ as follows:
\begin{equation}
\begin{split}
cp = \text{"\#Question:"} + C(S) + C(Q) +\\ \text{"\#Document:"} +  C(D) \\+ \text{"\#Let's think step by step"}
\end{split}
\end{equation}

Similarly, we define the output format, $co$, as:

\begin{equation}
    co = R + \text{"\#Answer:"} + A
\end{equation}

\paragraph{\texttt{BGQA}.} Firstly, we define the components of a data point in this dataset: facts ($F$), rules ($R$), and questions ($Q$). Therefore, our text prompt is defined as follows\footnote {\texttt{BGQA} provides a field \texttt{example} with all the variables of the dataset concatenated with descriptions. We use this field as text prompt.}:
\begin{equation}
\begin{split}
tp = F + R + Q
\end{split}
\end{equation}
This dataset also provides the CoT that leads to the answer. Therefore, we use that CoT as the expected output.

For code prompts, we follow the same approach as with the previous dataset. We define code prompts, $cp$, as follows:

\begin{equation}
tp = C(F) + C(R) + C(Q)
\end{equation}

with the output format ($co$) being:
\begin{equation}
co = C(cot)
\end{equation}

\paragraph{\texttt{ShARC}.} Firstly, we define the components of a data point in this dataset: question ($Q$), scenario ($S$), document ($D$), and conversation history ($H$). Then, the text prompt $tp$ is defined as follows:
\begin{equation}
\begin{split}
    tp = \text{"Question:"} + S + Q + \text{"Document:"} + D \\
    + \text{"Conversation history:"} + H \\
    + \text{"What is the answer to the question:"} + Q
\end{split}
\end{equation}
the output format is the answer label directly, which can be \textit{yes}, \textit{no}, or \textit{not enough information}.

Similarly to the other datasets, we defined code prompts $cp$ as follows:
\begin{equation}
\begin{split}
    tp = \text{"\#Question:"} + C(S) + C(Q) + \\
    + \text{"\#Document:"} + C(D) \\
    + \text{"\#Conversation history:"} + C(H) \\
    + \text{"\#What is the answer to the question:"} + C(Q)
\end{split}
\end{equation}
Lastly, the output format is the answer label directly, which in this case are \textit{True}, \textit{False}, or \textit{None}.

%% file: latex/sections/appendix/coding_features.tex
To generate code as close as possible to the NL text, we use a programming language based on a simplification of Python. We only use boolean variables or variables that contain lists of strings. Variables follow the snake case naming convention. We also employ \textit{if statements} to model conditional reasoning, but we do not use loops, functions, or classes. We create a code comment with the original NL text for each input sentence, and right after the code comment, we generate the code that represents the semantics of that sentence. However, we do not enforce the generated code to be a runnable script.

%% file: latex/sections/appendix/llm_setup.tex
The exact models we used are the following: gpt-3.5-16k-0613 for \texttt{CondQA} and \texttt{BGQA}. For \texttt{ShARC}, since the documents are shorter, we used GPT-3.5-0301 due to the lower costs. In both cases, we run the models through the Azure AI service. We also use Mixtral 8x7B with 4-bit quantization for all the datasets using one Nvidia A100 in our own server. Lastly, we use Mistral 7B v0.1 for \texttt{CondQA} and \texttt{BGQA}. However, this model yields very poor results on \texttt{ShARC}, so we use the \textit{instruct-v0.2} variant to be able to make a fair comparison between text and code prompts on this dataset using Mistral 7B. We use fp16 quantization for the Mistral 7B experiments and run them on our own server with one Nvidia A100.

All of our prompting methods are implemented using the Langchain library.\footnote{\url{https://github.com/langchain-ai/langchain}}
We set the decoding temperature to zero and use greedy sampling to make the outputs deterministic. 
For each experiment, we use a random sample from the training set as demonstrations. The LLM generating the code for code prompts is the same one as the one running the code to generate the final answer. We evaluate each model and prompt in the dev set of each dataset with two random seeds. Since the demonstrations are selected randomly, the seed determines them. The seed that yields the best performance on the dev set is then used for the final evaluation on the test set.

The number of demonstrations used to translate the documents into code is specified in \Cref{table:icl_code_transformation}. Note that this number differs from the number of demonstrations used to generate the answer, which is always three.

We use chain of thoughts (CoT) based on the provided annotations of the datasets. We do not use advanced CoT methods for text prompts because our aim is to quantify how much improvement we can get by transforming the natural language CoT into code syntax, and therefore, the natural language text and code must be \textit{as close as possible}. The use of advanced CoT methods would also be reflected in the code syntax, making the experimental setup more complicated without providing better insights.

The best random seeds found (and consequently used for the test set evaluation) are described in \Cref{table:seeds_code} and \Cref{table:seeds_text}.

\begin{table}[h]
    \centering
    \begin{tabular}{lccc}
    \toprule
         \textbf{Dataset} & \textbf{GPT} & \textbf{Mixtral} & \textbf{Mistral} \\
         \midrule
         \texttt{CondQA}  &  4 & 4 & 4 \\
         \texttt{ShARC}   &  5  & 4 & 4 \\
         \texttt{BGQA-1}  &  4 & 3 & 3 \\
         \texttt{BGQA-2}  &  4 & 3 & 4 \\
         \texttt{BGQA-3}  &  4 & 3 & 4 \\
         \bottomrule
    \end{tabular}
    \caption{Number of demonstrations for code translations. Note this is not the number of demonstrations to generate the answer.}
    \label{table:icl_code_transformation}
\end{table}

\begin{table}[h]
    \centering
    \begin{tabular}{lccc}
    \toprule
         \textbf{Dataset} & \textbf{GPT} & \textbf{Mixtral} & \textbf{Mistral} \\
         \midrule
         \texttt{CondQA}  &  0 & 0 & 0 \\
         \texttt{ShARC}   &  0 & 1 & 1 \\
         \texttt{BGQA-1}  & 1 & 0 & 1 \\
         \texttt{BGQA-2}  & 1  & 0 & 0\\
         \texttt{BGQA-3}  & 0  & 1 &  0\\
         \bottomrule
    \end{tabular}
    \caption{Best seeds for code prompts}
    \label{table:seeds_code}
\end{table}

\begin{table}[h]
    \centering
    \begin{tabular}{lccc}
    \toprule
         \textbf{Dataset} & \textbf{GPT} & \textbf{Mixtral} & \textbf{Mistral} \\
         \midrule
         \texttt{CondQA}  & 0  & 1 & 0\\
         \texttt{ShARC}   &  0  & 0 & 0\\
         \texttt{BGQA-1}  &  1 & 0 &  1\\
         \texttt{BGQA-2}  &  0 & 1 & 1\\
         \texttt{BGQA-3}  &  0 & 1 & 0 \\
         \bottomrule
    \end{tabular}
    \caption{Best seeds for text prompts}
    \label{table:seeds_text}
\end{table}

%% file: latex/sections/appendix/costs.tex
Running a data instance from ConditionalQA with gpt-3.5-16k-0613 using code prompts costs \$0.04 while with text prompts \$0.01. On BoardgameQA-depth 3 (i.e., the partition with the most expensive prompts), with the same model, the costs per question are \$0.02 and \$0.03 for text and code prompts, respectively.  Lastly, on ShaRC, using gpt-3.5-0301, the costs per question are \$0.0006 and \$0.005 for text and code prompts, respectively.

%% file: latex/sections/appendix/codellm.tex
Although our work focuses on text+code LLMs because they are the only type of LLMs whose intended use includes natural language and coding tasks, we conduct a small experiment on \texttt{Code Llama} \citep{roziere2023code}, a code-only LLM. It is important to note that their authors advise against using this model on natural language tasks because their intended use is in code generation tasks only. \Cref{table:codellama} shows the results of \texttt{Code Llama} on our datasets. Firstly, we can observe that code prompts perform significantly worse than text prompts on \texttt{CondQA} and \texttt{ShARC} despite being a code LLM. We can attribute this to the nature of these datasets and the intended use of the model. These datasets require a strong comprehension of natural language documents and dialogues and answering natural language questions about them. This is far from the intended use of the model (i.e., generating code). Furthermore, \texttt{CondQA} requires generating a natural language answer that is a span of the document. The use of code to generate a natural language span of a document is also far from the fine-tuning tasks of this model. This would explain why the code representation is worse than the text representation. It is particularly interesting to see the results on ShARC. After manually inspecting the outputs, we observe that \texttt{Code Llama} can successfully generate the code corresponding to the natural language input. However, when it is prompted with such code and the question variable, the model does not generate the value of the variable (i.e.,  \texttt{true}, \texttt{false}, or \texttt{none}). Instead, it generates \texttt{\textbackslash n}. The reasons behind this remain unclear and would require further investigation, which is out of the scope of this paper.

However, we observe a different behavior on \texttt{BGQA}. In this dataset, code prompts outperform text prompts. We attribute this to the high alignment with the first-order logic of this dataset, which makes it closer to the intended use of the model.

Nevertheless, it is important to note that these results are not intended to be comprehensive enough to conclude that code LLMs or \texttt{Code Llama} can or cannot solve natural language tasks, which is out of the scope of this work. Instead, they simply seem to confirm the warnings of the authors of \texttt{Code Llama}, i.e., this model is not intended for natural language tasks.

\begin{table}[h]
\centering
\begin{tabular}{lcc}
\toprule
\textbf{Model}  & \textbf{Text Prompt} & \textbf{Code Prompt} \\
\midrule
CondQA & 31.58                      & 26.34                      \\
ShARC  & 58.33                      & 18.62                      \\
BGQA1  & 44.38                      & 44.78                      \\
BGQA2  & 44.59                      & 49.41                      \\
BGQA3  & 40.88                      & 46.44                      \\
\bottomrule
\end{tabular}
\caption{Text and code prompts results in \texttt{Code Llama 7B - Instruct} with one demonstration.}
\label{table:codellama}
\end{table}

%% file: latex/sections/appendix/textllm.tex
As briefly mentioned in \Cref{sec:models}, text-only LLMs are not expected to perform well with code prompting and should not be used for this as they are not explicitly trained on code. For example, on the MBPP coding benchmark \citep{austin2021program}, LLaMA 2 scores 26.1\% , while Mistral 7B (a text+code LLM) achieves 47.5\% and code-llama 2 7B, achieves 52.5\% \citep{jiang2023mistral}. \Cref{table:llama2} further proves our claim. LLaMA 2 7B-Chat \citep{touvron2023llama2openfoundation} with code prompting never outperforms text prompting due to its lack of code understanding.

\begin{table}[h]
\centering
\begin{tabular}{@{}lcccc@{}}
\toprule
\textbf{Prompt}             & \textbf{CQA} & \textbf{ShARC} & \textbf{BG\textsubscript{1} }&\textbf{ BG\textsubscript{2}} \\ \midrule
Text  & \textbf{29.79}  & \textbf{46.16} & \textbf{51.85}  & \textbf{39.23}  \\
Code   & 21.32  & 24.74 & 45.16  & 37.66  \\ \bottomrule
\end{tabular}
\caption{LLaMA 2 7B Chat Results. Code prompts cannot perfom well on text-only LLMs.}
\label{table:llama2}
\end{table}

%% file: latex/sections/appendix/phi2_results.tex
We have shown the effectiveness of code prompting in the most popular sizes of LLMs in \cref{table:main_results} from \cref{sec:codevstext}. However, it is becoming increasingly popular the development of small language models (sLMs) due to their cheaper inferece cost and higher token thoughput \citep{gunasekar2023textbooks}. Therefore, we have conducted a preliminary experiment with Phi-2\footnote{\url{https://huggingface.co/microsoft/phi-2}}, a text+code model of 2.7B parameters on \texttt{BGQA-1} to show that our prompting methodology also holds in sLMs. As we can show on \cref{table:phi2_results}, code prompting yields a remarkable performance boost of 15 points. However, due to the limited context window of Phi-2, it is not straightforward to conduct in-context learning on our other datasets.

\input{latex/new_tables/phi2_results}

%% file: latex/new_tables/phi2_results.tex
\begin{table}[t]
\begin{center}
\begin{tabular}{llccccccc}
\toprule
\textbf{Prompt}  & \textbf{BGQA-1} \\
\midrule
Text & $ 33.20 \pm 1.42$ \\
Code & \textbf{48.32} $\pm$ \textbf{1.65}  \\

\bottomrule
\end{tabular}
\end{center}
\caption{Comparison of text prompt and code prompts with Phi-2 on the validation set. Metric: F1 score. One demonstration per class is provided.}
\label{table:phi2_results}
\end{table}

%% file: latex/sections/appendix/ablations_local_llms.tex
\Cref{table:mistral_ablations} shows that all ablations to code prompting in Mistral yield significant performance drops similar to those observed in GPT 3.5

\begin{table}[h]
\centering
\begin{tabular}{@{}lccc@{}}
\toprule
\textbf{Prompt} & \textbf{BG\textsubscript{1} }&\textbf{ BG\textsubscript{2}} & \textbf{ BG\textsubscript{3}} \\
\midrule
Atomic St.         & -4.29          & -8.09          & -2.19          \\
NLCode             & -2.7           & -4.36          & -5.97          \\
Anonym. Code       & -0.45          & -7.17          & -4.52          \\
Rnd Code           & -4.27          & -11.16         & -13.93         \\
- Comments         & -0.79          & -9.88          & -29.21      \\
\bottomrule
\end{tabular}
\caption{Performance drop w.r.t. Code Prompting on Mistral.}
\label{table:mistral_ablations}
\end{table}

\Cref{table:mixtral_ablations} shows that most ablations to code prompting in Mixtral also yield significant performance drops similar to those observed in GPT 3.5, except on \texttt{BGQA-1}, where text prompts outperform code prompts. Therefore, it is expected that the atomic statements and natural language code ablations improve performance.
\begin{table}[h]
\centering
\begin{tabular}{@{}lccc@{}}
\toprule
\textbf{Prompt} & \textbf{BG\textsubscript{1} }&\textbf{ BG\textsubscript{2}} & \textbf{ BG\textsubscript{3}} \\ \midrule
Atomic St.         & 1.47           & -4.01          & -7.61          \\
NLCode             & 0.9            & -2.82          & -3.18       \\
Anonym. Code       & 0.21           & -16.59         & 1.25           \\
Rnd Code           & -15.88         & -9.56          & -11.31         \\
- Comments         & -12.81         & -7.14          & 5.38           \\ \bottomrule
\end{tabular}
\caption{Performance drop w.r.t. Code Prompting on Mixtral.}
\label{table:mixtral_ablations}
\end{table}

%% file: latex/sections/appendix/demonstrations_local_llms.tex
We conduct experiments with Mistral and Mixtral on \texttt{BGQA} because it is the dataset where we clearly see a difference between text and code prompts with different numbers of demonstrations. \Cref{table:demonstrations_mistral} shows that Mistral achieves the best results with just one demonstration for both prompting methods. We attribute this to the long length and complexity of the demonstrations\footnote{We cannot even provide three demonstrations for \texttt{BGQA-2} and \texttt{3} due to the length of each demonstration.}, which can confuse LLMs, especially those that are small, which is the case for Mistral. \citet{li2024long} shows that Mistral dramatically decreases its performance as the number of input tokens increases. However, very large LMs are more resilient to the input length. Text prompts on Mixtral achieve the best results with at least two demonstrations, while code prompts only achieve significantly better results with more than one demonstration in \texttt{BGQA-3}.

\begin{table}[]
\centering
\setlength{\tabcolsep}{4pt}
\begin{tabular}{@{}lllccc@{}}
\toprule
\textbf{Model}                    & \textbf{Prompt}             & \textbf{\# Dem.} & \textbf{BG\textsubscript{1} }&\textbf{ BG\textsubscript{2}} & \textbf{ BG\textsubscript{3}} \\ \midrule
\multirow{6}{*}{Mistral} & \multirow{3}{*}{Text} & 1       & \textbf{48.26} & 45.90  & \textbf{48.63} \\
                         &                       & 2       & 45.86 & \textbf{48.82} & 47.92 \\
                         &                       & 3       & 44.25 & N.A. & N.A. \\\cmidrule(l){2-6}
                         & \multirow{3}{*}{Code} & 1       & 51.54 & \textbf{55.28} & \textbf{52.83} \\
                         &                       & 2       & \textbf{51.85} & 50.51 & 48.66 \\
                         &                       & 3       & 39.58 & N.A. &   N.A.    \\ \midrule
\multirow{6}{*}{Mixtral} & \multirow{3}{*}{Text} & 1       & 57.06 & 36.71 & 30.8  \\
                         &                       & 2       & \textbf{65.49} & \textbf{39.83} & \textbf{33.23} \\
                         &                       & 3       & 63.73 &  N.A.     &   N.A.    \\  \cmidrule(l){2-6}
                         & \multirow{3}{*}{Code} & 1       & \textbf{61.85} & 44.93 & 35.22 \\
                         &                       & 2       & 58.44 & 45.70  & \textbf{39.43} \\
                         &                       & 3       & 61.3  &  N.A.     &    N.A.   \\ \bottomrule
\end{tabular}
\caption{Text and code prompting performance on Mistral and Mixtral under a different number of demonstrations. Significantly best results in bold.}
\label{table:demonstrations_mistral}
\end{table}

%% file: latex/sections/appendix/variable-tracking.tex
\paragraph{Extracting key entities in BoardgameQA.} This dataset provides a list of ``\textit{facts,}'' which are short and concise sentences describing the state of a key entity. Therefore, we use them without alterations as the key entities to ask for.

\paragraph{Extracting key entities in ConditionalQA.} This dataset provides a scenario describing the background information of the person posing the answer. Since this scenario is a free-form text, we follow \citep{min-etal-2023-factscore} to extract \textit{atomic statements} and use them as the key entities to ask for.

\paragraph{Code Prompting variables}. To probe the variable tracking abilities of code prompts, we use the variables defined in the ``\textit{facts}'' and ``\textit{scenario}'' of BoardgameQA and ConditionalQA, respectively.

\paragraph{Probing memory at different steps in the Chain-of-Thought.} Inspired by \citet{lanham2023measuring}, we truncate the Chain-of-Thought (CoT) at different completion states and probe the memory of the model. To break down the CoT, we split it by the character ``\textbackslash n'', which usually represents the end of a reasoning step. This is possible because our in-context learning demonstrations follow this format.

\paragraph{Number of probes.} For each dataset instance, we run $num\_facts \times num\_steps\_cot$ probes, which makes this experiment very costly. Thus, we aim to maximize the number of instances probed while keeping the costs down. To do so, we use a sample of 50 instances for each dataset partition of BoardgameQA, except for Board3, where we used 20 instances ($\approx 700$ probes) because of the cost of the experiment. Due to the length of the demonstrations of ConditionalQA and its impact on the costs, we sample five facts and three partial CoTs for each instance, yielding an upper-bound of 15 probes per instance, and run the probes for 30 instances for each dataset partition (i.e., correct and incorrect instances).

\paragraph{Prompt Probes.}
In all cases, we follow the following format: \textit{Sys. Prompt; ICL Demonstrations; Input Instance; Partial CoT; \textbf{Probe}}.

The probe for text and code prompts in BoardgameQA is: ``Now, I want to ask you about the value of some key entities you used. Your answers must be `yes`, `no`, or `unknown`. It is very important that you only write one word. Is it true that \{fact\}?''

The probe for text prompts in ConditionalQA is: ``Now, I want to ask you about the value of some key entities you used. Your answers must be ``True'', ``False'', ``unknown'', or a string. It is very important that you only write the exact value. From the speaker perspective, is it true that \{fact\}?''

The probe for code prompts in ConditionalQA is: ``Now, I want to ask you about the value of some key entities you used. Your answers must be ``True'', ``False'', ``unknown'', or a string. It is very important that you only write the exact value. What is the value of the variable \{var\}?'' A real example is provided in \Cref{table:variable_tracking_example}.

%% file: latex/sections/appendix/memory_errors_local_llms.tex
Analyzing the memory errors in Mixtral and Mistral becomes much more challenging than in GPT 3.5 due to the lower performance of the models. \Cref{table:mixtral_memory} shows that, in general, Mixtral makes significantly more memory errors than GPT 3.5. On \texttt{BGQA-1}, we observed more memory errors on the questions where the model fails than when the model returns the correct answer, as expected. We further see that code prompts make fewer memory errors than text prompts, confirming the results of GPT 3.5. However, the margins are much narrower than in GPT 3.5.

On \texttt{BGQA-2} and \texttt{3}, the model surprisingly makes fewer memory errors on the wrong answer partition. However, in both cases, the percentage is so high that makes the test unreliable, as we observe in CondQA in GPT 3.5. In the specific case of Mistral, we were not able to run this experiment successfully due to the model not understanding the task. We believe our experiment setup for memory error analysis only works well on very high-performing models such as GPT. To analyze them on smaller models such as Mistral and Mixtral, we would need to devise other types of experiments, which is out of the scope of this work. We encourage further research on entity tracking methods to analyze LLMs and leave this as future work.

\begin{table}[t]
\centering
\begin{tabular}{@{}lcccc@{}}
\toprule
      & \multicolumn{2}{c}{\textbf{Correct Ans.}} & \multicolumn{2}{c}{\textbf{Incorrect Ans.}} \\ 
      & \textbf{Text}            & \textbf{Code}           & \textbf{Text}             & \textbf{Code}            \\\midrule
\texttt{BGQA-1} & 58.5            & 54.0             & 60.4             & 55.6            \\
\texttt{BGQA-2} & 70.8            & 77.6           & 66.0               & 80.0              \\
\texttt{BGQA-3} & 65.2            & 80.4           & 62.8             & 94.4            \\ \bottomrule
\end{tabular}
\caption{Percentage of memory errors in Mixtral.}
\label{table:mixtral_memory}
\end{table}

%% file: latex/sections/appendix/confusion_matrix.tex
\begin{figure*}[t]
\centering
\includegraphics[]{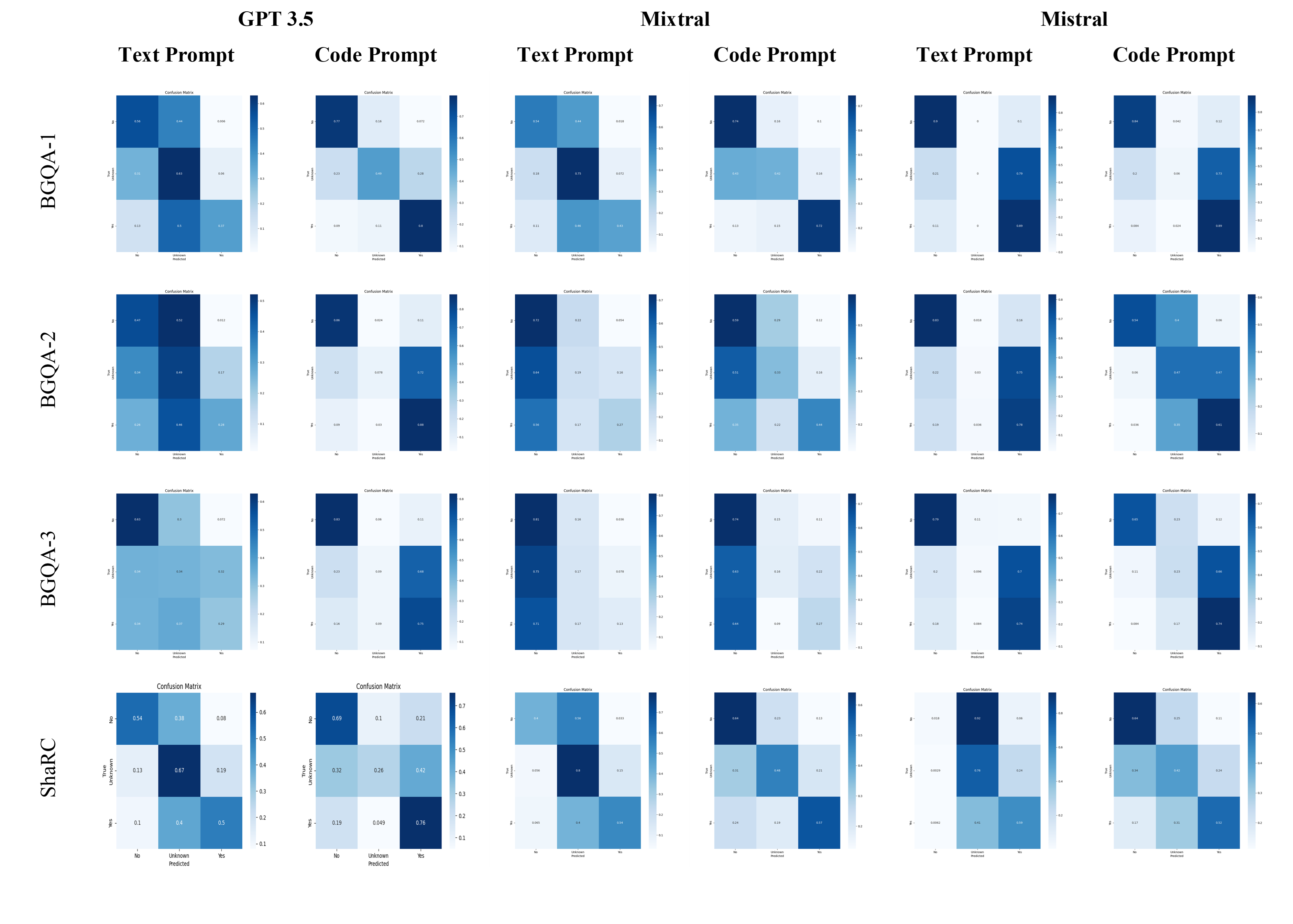}
\caption{Confusion matrices of text and code prompts for each model on all datasets.}
\label{fig:confusion_matrix}
\end{figure*}

\Cref{fig:confusion_matrix} shows the confusion matrices of all our models using text and code prompts for all the datasets except \texttt{CondQA}. We cannot include this one because it is a span-extraction task, not a classification task.

%% file: latex/sections/appendix/atomic_statements.tex
Original sentence: <p>Applying for the legal right to deal with someone's property, money and possessions (their estate) when they die is called applying for probate.</p>
Atomic statements: Applying for the legal right is a process. The process is called 'applying for probate'. The legal right is to deal with someone's property, money, and possessions. The someone is a person who has died. The property, money, and possessions are collectively called the 'estate'.

%% file: latex/sections/appendix/code_ablations.tex
An example of a back-translated code into natural language is provided in \Cref{table:back_translation_example}. We can observe in both examples that the resulting natural language (NL) text is extremely similar to the original code. In addition, in the second example (\texttt{BGQA}), \textit{Rule2} is much simpler after the back-translation than its original description in NL.

\Cref{table:code_ablations_example} shows examples of the multiple code ablations we conducted in \Cref{sec:ablation}. Random code replaces the code with a piece of code from another data point. In this way, the semantics of the text and code mismatch while we keep the code syntactically correct.

%% file: latex/sections/appendix/code_ablations_tables.tex
\begin{table*}
    \centering
    \begin{tabular}{p{0.12\linewidth} | p{0.80\linewidth} }
    \toprule
        \textbf{Type} & \textbf{Text}  \\
        \midrule         
        Code & 
        \# <p>You can apply to become the estate’s administrator if you are 18 or over and you are the most ‘entitled’ inheritor of the deceased’s estate. This is usually the deceased’s closest living relative.</p>
         
        \textbf{\texttt{if applicant\_age >= 18 and entitled\_inheritor and closest\_relative:}}
	   
        \quad   \textbf{\texttt{can\_apply\_estate\_administrator = True}} \\
        Code $\rightarrow$ NL  & <p>You can apply to become the estate’s administrator if you are 18 or over and you are the most ‘entitled’ inheritor of the deceased’s estate. This is usually the deceased’s closest living relative.</p>
         
        \textbf{if you are 18 or over and you are the most entitled inheritor of the deceased's estate and you are the closest living relative, you can apply to become the estate's administrator}\\
        \midrule
        Code & \# Rule2: Be careful when something removes from the board one of the pieces of the dog and also becomes an enemy of the catfish because in this case it will surely not burn the warehouse of the mosquito (this may or may not be problematic)

        \textbf{rule2(something) = remove(something, piece\_of(dog)) \& enemy(something, catfish) => not burn(something, warehouse\_of(mosquito))}
        \\
        Code $\rightarrow$ NL  & \textbf{Rule2: If something removes from the board one of the pieces of the dog and also becomes an enemy of the catfish, then it does not burn the warehouse of the mosquito} \\

         \bottomrule
    \end{tabular}
    \caption{Example of a back-translation ${\mathbb{NL} \rightarrow \mathbb{C}}$ in ConditionalQA and BGQA-3. Text in bold represents the main modification.}
    \label{table:back_translation_example}
\end{table*}

\begin{table*}
    \centering
    \begin{tabular}{p{0.12\linewidth} | p{0.80\linewidth} }
    \toprule
        \textbf{Type} & \textbf{Text}  \\
        \midrule 
        Original Code & \# <p>To be eligible you must have left your country and be unable to go back because you fear persecution.</p>
        
        \textbf{\texttt{if left\_country\_and\_fear\_persecution:}}
        
        \quad   \textbf{\texttt{eligible\_for\_asylum = True}}
        \\
        
        \midrule
        Anonymous Code  & 
        \# <p>To be eligible you must have left your country and be unable to go back because you fear persecution.</p>
        
        \textbf{\texttt{if var\_1}}
        
        \quad   \textbf{\texttt{var\_2 = True}}
        \\
        \midrule
        Random Code &  \# <p>To be eligible you must have left your country and be unable to go back because you fear persecution.</p>
        
        \textbf{\texttt{if value\_of\_property\_gone\_down\_by\_more\_than\_50:}}

        \quad   \textbf{\texttt{eligible\_to\_claim = True}}

        \quad   \textbf{\texttt{getting\_housing\_benefit = True}}
\\
        
         \bottomrule
    \end{tabular}
    \caption{Examples code ablations.}
    \label{table:code_ablations_example}
\end{table*}

%% file: latex/sections/appendix/variable_tracking_example.tex
\begin{table*}
    \centering
    \begin{tabular}{p{0.1\linewidth} | p{0.1\linewidth} | p{0.71\linewidth}}
    \toprule
         \textbf{Section} & \textbf{Role} & \textbf{Message} \\
         \midrule
         Problem instance & Human & \textbf{Question}: My brother and his wife are in prison for carrying out a large fraud scheme. Their 7 and 8 year old \underline{children have been living with me for the last 4 years}. I want to become their Special Guardian to look after them permanently. How long will it be before I hear back from the court?
         
         \textbf{Document}: <h1>What is a special guardian</h1> <p>You can apply to be a child’s special guardian when they cannot live with their birth parents and adoption is not right for them.</p> ...
         
         Answers can be "yes" or "no".  Let's think step by step:\\
         Partial CoT & AI &  <p>Within 10 days of receiving your application the court will send you a case number and a date for a meeting to set out:</p>\textbackslash n \\
         \midrule
         Probe & Human & Now, I want to ask you about the value of some key entities you used. Your answers must be `True`, `False`, `unknown`, or a string. It is very important that you only write the exact value. From the speaker perspective, is it true that \underline{the children have been living with me for the last 4 years}?
 \\
         Probe & AI & True \\
         \bottomrule
    \end{tabular}
    \caption{Variable Tracking Example. Underlined text represents the variable to probe. Partial CoT is not the complete answer. The generation was stopped, and only the first step was used in this probe.}
    \label{table:variable_tracking_example}
\end{table*}

%% file: latex/sections/appendix/prompt_examples.tex
\input{latex/sections/appendix/prompt_examples/condqa/text_prompt}
\input{latex/sections/appendix/prompt_examples/condqa/code_prompt}
\input{latex/sections/appendix/prompt_examples/bgqa/text_prompt}
\input{latex/sections/appendix/prompt_examples/bgqa/code_prompt}
\input{latex/sections/appendix/prompt_examples/sharc/text_prompt}
\input{latex/sections/appendix/prompt_examples/sharc/code_prompt}

%% file: latex/sections/appendix/prompt_examples/condqa/text_prompt.tex
\begin{table*}
    \centering
    \scalebox{0.80}{
    \begin{tabular}{p{\linewidth} }
    \toprule
        System: You are a helpful assistant that answers questions given a document. Answers must be a short span of the document. You have to extract the span from the document. Do not write anything else. I will give you some examples first.

        ICL Demonstrations...
        
    Human: Question: My brother and his wife are in prison for carrying out a large fraud scheme. Their 7 and 8 year old children have been living with me for the last 4 years. I want to become their Special Guardian to look after them permanently. How long will it be before I hear back from the court?
    
Document: <h1>What is a special guardian</h1>

<p>You can apply to be a child’s special guardian when they cannot live with their birth parents and adoption is not right for them.</p>

<p>You’ll be responsible for looking after the child until they’re 18 (unless the court takes your responsibility away earlier).</p>

<p>You’ll make all day to day decisions about the child, for example schooling and medical treatment. You do not have to discuss these decisions with the birth parents.</p>

<p>You’ll need to get the consent of everyone who has parental responsibility for the child before you make some important decisions, for example:</p>

<li>changing the child’s surname</li>

<li>putting the child up for adoption</li>

<li>taking the child abroad for more than 3 months</li>

<li>the child having surgery for reasons other than improving health, such as circumcision, sterilisation or cosmetic surgery</li>

<p>If you cannot get consent, you can ask the court to decide. Use the form ‘Make an application in existing court proceedings related to children’ (form C2).</p>

<h1>After you apply</h1>

<p>Within 10 days of receiving your application the court will send you a case number and a date for a meeting to set out:</p>

<li>a timetable for your case</li>

<li>how it will be dealt with</li>

<p>This meeting is called a ‘first directions hearing’.</p>

<p>You must go to all hearings you’re told to unless the court excuses you. If you’re not able to go, contact the court office.</p>
Answers must be a short span of the document. You have to extract the span from the document. Do not write anything else. Let's think step by step:
        \\
         \bottomrule
    \end{tabular}
    }
    \caption{Text prompt Example for ConditionalQA}
    \label{table:condqa_text_prompt_example}
\end{table*}

%% file: latex/sections/appendix/prompt_examples/condqa/code_prompt.tex
\begin{table*}
    \centering
    \scalebox{0.65}{
    \begin{tabular}{p{\linewidth} }
    \toprule
        System: You are a helpful assistant. Your task is to process a pseudo-code that describes a question and a document. You need to reason using that document and the comments to return the answers. Answers must be a short span of the document. You have to extract the span from the code comments. Do not write anything else. I will give you some examples first.

        ICL Demonstrations...
        
        Human: \# Question: My brother and his wife are in prison for carrying out a large fraud scheme. Their 7 and 8 year old children have been living with me for the last 4 years. I want to become their Special Guardian to look after them permanently. How long will it be before I hear back from the court?
        
        maximum\_redundancy\_pay = 16320
        
        housing\_standards\_and\_procedures\_in\_Northern\_Ireland = True
        
        ensure\_vehicle\_taxed\_in\_UK = True
        immigration\_advisers\_can\_help\_with\_representation\_at\_tribunal = True
        
        supply\_protective\_clothing\_and\_equipment = True
        
        CBT\_required\_for\_moped\_and\_motorcycle = True
        
        court\_response\_time = None \# This is the variable that answers the question

        \# <h1>What is a special guardian</h1>
        
\# <p>You can apply to be a child’s special guardian when they cannot live with their birth parents and adoption is not right for them.</p>

if attorneys\_appointed\_jointly:

    all\_attorneys\_must\_agree\_to\_make\_decision = True

disability\_or\_severe\_disability\_element\_of\_working\_tax\_credit = True

mugging\_without\_physical\_harm\_emergency = True

\# <p>You’ll be responsible for looking after the child until they’re 18 (unless the court takes your responsibility away earlier).</p>

work\_temporarily\_for\_hirer = True

\# <p>You’ll make all day to day decisions about the child, for example schooling and medical treatment. You do not have to discuss these decisions with the birth parents.</p>

accounts\_and\_tax\_returns\_cover\_financial\_year = "1 June to 31 May"

employer\_operating\_PAYE = True

\# <p>You’ll need to get the consent of everyone who has parental responsibility for the child before you make some important decisions, for example:</p>

\# <li>changing the child’s surname</li>

\# <li>putting the child up for adoption</li>

\# <li>taking the child abroad for more than 3 months</li>

\# <li>the child having surgery for reasons other than improving health, such as circumcision, sterilisation or cosmetic surgery</li>

managed\_by\_fit\_and\_proper\_persons = True

check\_court\_order\_for\_authorization = True

considering\_fostering = True

if not\_connected\_to\_mains\_sewer:

    septic\_tank\_used = True

can\_claim\_tax\_relief\_if\_taxed\_twice = True

extra\_support\_for\_disability = True

if operator\_of\_septic\_tank\_or\_treatment\_plant:

    follow\_general\_binding\_rules = True

\# <p>If you cannot get consent, you can ask the court to decide. Use the form ‘Make an application in existing court proceedings related to children’ (form C2).</p>

appeals\_decision\_time = "several months"

if worker and informal\_resolution\_not\_satisfactory:

    formal\_grievance\_complaint\_possible = True

    time\_limit\_for\_backdating\_claims\_services = 6

\# <h1>After you apply</h1>

\# <p>Within 10 days of receiving your application the court will send you a case number and a date for a meeting to set out:</p>

\# <li>a timetable for your case</li>

\# <li>how it will be dealt with</li>

\# <p>This meeting is called a ‘first directions hearing’.</p>

committee\_recommendations\_go\_to\_Prime\_Minister = True

check\_adviser\_registration = True

meet\_manning\_levels = True

recognised\_as\_charity\_or\_CASC = True

apply\_for\_visa\_for\_other\_reasons = True

debt\_paid\_off = True

if special\_educational\_needs\_and\_disabilities:

    affects\_behaviour\_or\_socialisation = True

\# <p>You must go to all hearings you’re told to unless the court excuses you. If you’re not able to go, contact the court office.</p>

payslip\_can\_include\_tax\_code = True

VAT\_zero\_rate = 0

gas\_equipment\_installed\_and\_maintained\_by\_Gas\_Safe\_registered\_engineer = True

\# Question: My brother and his wife are in prison for carrying out a large fraud scheme. Their 7 and 8 year old children have been living with me for the last 4 years. I want to become their Special Guardian to look after them permanently. How long will it be before I hear back from the court?

\# Answers must be a short span of the document. You have to extract the span from the code comments. Do not write anything else.

\# Let's think step by step:       
        \\
         \bottomrule
    \end{tabular}
    }
    \caption{Code Prompt Example for ConditionalQA}
    \label{table:condqa_cond_prompt_example}
\end{table*}

%% file: latex/sections/appendix/prompt_examples/bgqa/text_prompt.tex
\begin{table*}
    \centering
    \begin{tabular}{p{0.95\linewidth} }
    \toprule
        System: You are a question-answering system that solves the problem of reasoning with contradictory information guided by preferences over sources of information. You must explain your answers step by step.

        ICL Demonstrations ...

        Human: A few players are playing a boardgame

The current state of the game is as follows

The amberjack struggles to find food

And the rules of the game are as follows

Rule1: If the amberjack has difficulty to find food, then the amberjack removes from the board one of the pieces of the carp

Based on the game state and the rules and preferences, does the amberjack remove from the board one of the pieces of the carp?

        AI:
        \\
         \bottomrule
    \end{tabular}
    \caption{Text prompt Example for BGQA-1}
    \label{table:bgqa_text_prompt_example}
\end{table*}

%% file: latex/sections/appendix/prompt_examples/bgqa/code_prompt.tex
\begin{table*}
    \centering
    \begin{tabular}{p{0.95\linewidth} }
    \toprule
        System: You are a large language model of code that can interpret code. You are given a pseudo-code that resembles to first-order logic that models some scenario. You will be given a question and you have to answer it step by step. You can use a rule if and only if you know the antecedent of the rule.

        ICL Demonstrations
        
        Human: \# A few players are playing a boardgame

\# The rules of the game are as follows

\# Rule1: If the amberjack has difficulty to find food, then the amberjack removes from the board one of the pieces of the carp.

rule1() = difficulty\_finding\_food(amberjack) => remove\_piece(amberjack, carp)

\# The current state of the game is as follows

\# The amberjack struggles to find food.

difficulty\_finding\_food(amberjack) = True

\# Based on the game state and the rules and preferences, does the amberjack remove from the board one of the pieces of the carp?

question = remove\_piece(amberjack, carp)

AI:
        \\
         \bottomrule
    \end{tabular}
    \caption{Code prompt Example for BGQA-1}
    \label{table:bgqa_code_prompt_example}
\end{table*}

%% file: latex/sections/appendix/prompt_examples/sharc/text_prompt.tex
\begin{table*}
    \centering
    \scalebox{0.95}{
    \begin{tabular}{p{\linewidth} }
    \toprule
        System: You are a question answering system that answers questions given a document and a conversation history. The conversation history gives information about the background of the person posing the question. You must answer `yes`, `no`, or `not enough information` to the question and nothing else.

        ICL Demonstrations...
        
    Human: Question: The item is not equipment for audio books or newspapers, and I'm not selling lifeboats or anything related to that. It's for medicine and medicinal ingredients. Can I apply zero VAT to this item?
Document: 

\#\# Items that qualify for the zero rate

You may be able to apply zero VAT when you sell the following to an eligible charity:

* equipment for making ‘talking’ books and newspapers

* lifeboats and associated equipment, including fuel

* medicine or ingredients for medicine

* resuscitation training models

Conversation history:

Q: Is it equipment for making ‘talking’ books and newspapers?

A: No

Q: Are you selling lifeboats and associated equipment, including fuel?

A: No

Q: Are you selling medicine or ingredients for medicine?

A: Yes

What is the answer to the question: Can I apply zero VAT to this item? You must answer `yes`, `no`, or `not enough information` to the question and nothing else.

AI:
        \\
         \bottomrule
    \end{tabular}
    }
    \caption{Text prompt Example for ShARC.}
    \label{table:sharc_text_prompt_example}
\end{table*}

%% file: latex/sections/appendix/prompt_examples/sharc/code_prompt.tex
\begin{table*}
    \centering
    \scalebox{0.95}{
    \begin{tabular}{p{\linewidth} }
    \toprule
        System: You are a question-answering system that answers questions based on a document, and conversation history. The text is pseudo-code that models the document and conversation history. You must run the code and update the value of the variable that answers the question. The values can be True, False, or None.

        ICL Demonstrations...
        
    Human:
    
    \# Question:
\# The item is not equipment for audio books or newspapers, and I'm not selling lifeboats or anything related to that. It's for medicine and medicinal ingredients. Can I apply zero VAT to this item?

equipment\_for\_audio\_books\_or\_newspapers = False

selling\_lifeboats\_or\_related\_equipment = False

selling\_medicine\_or\_ingredients\_for\_medicine = True

can\_apply\_zero\_VAT = None \# This is the variable that answers the question.

\# Other variables needed for the document:

\# Document:

\#\# Items that qualify for the zero rate

\# You may be able to apply zero VAT when you sell the following to an eligible charity:

\# * equipment for making ‘talking’ books and newspapers

if equipment\_for\_audio\_books\_or\_newspapers:

    can\_apply\_zero\_VAT = False

\# * lifeboats and associated equipment, including fuel

if selling\_lifeboats\_or\_related\_equipment:

    can\_apply\_zero\_VAT = False

\# * medicine or ingredients for medicine

if selling\_medicine\_or\_ingredients\_for\_medicine:

    can\_apply\_zero\_VAT = True

\# * resuscitation training models

resuscitation\_training\_models = None

can\_apply\_zero\_VAT = 

AI: 
        \\
         \bottomrule
    \end{tabular}
    }
    \caption{Code prompt Example for ShARC.}
    \label{table:sharc_code_prompt_example}
\end{table*}